  \documentclass[final,5p,times]{elsarticle}

\usepackage{graphicx}
\usepackage{lscape}
\usepackage{verbatim}
\usepackage{algorithmic}
\usepackage{algorithm2e}
\usepackage{amsmath}
\usepackage{amssymb}

\usepackage{color,soul}

\usepackage{xcolor}
\usepackage{subfigure}

\usepackage{tabularx,ragged2e,booktabs,caption,array,multirow,multicol}
\usepackage{csquotes}
\usepackage{mathtools} 

\usepackage{lineno}
\usepackage{amsthm}  
\usepackage{listings}
\usepackage{amssymb}
\usepackage{latexsym}
\usepackage{epsfig}
\usepackage{float}
\usepackage{xspace}
\usepackage{algorithmwh}
 \usepackage{float}

   \graphicspath{{../pdf/}{../jpeg/}}
  \DeclareGraphicsExtensions{.pdf,.jpeg,.png}

\journal{ }

\begin{document}

\begin{frontmatter}

\title{Co-evolutionary multi-task learning for dynamic time series prediction}

 \author[label1, label2 ]{Rohitash Chandra} 
 \author[label3]{Yew-Soon Ong} 
\author[label3]{Chi-Keong Goh }

\address[label1]{  Centre for Translational Data Science, The University of 
Sydney, Sydney, NSW 2006, Australia \\}

\address[label2]{  School of Geosciences, The University of 
Sydney, Sydney, NSW 2006, Australia \\}

\address[label3]{ Rolls Royce @NTU Corp Lab,  Nanyang Technological 
University, 42 Nanyang View, Singapore  \\}

\begin{abstract}

Time series prediction typically 
consists of a data reconstruction phase where the time series is broken into 
overlapping windows known as the timespan. The size of the timespan  can be 
seen as a way of determining the extent of past information required for   an 
effective prediction.    
In certain applications such as the prediction of wind-intensity of storms and cyclones,  prediction 
models need to be dynamic in accommodating different values of the timespan. 
These applications require robust prediction  as soon as the event takes place.  
We identify a new 
category of problem called   dynamic time series 
prediction that requires a  model  to give prediction when 
presented with varying lengths  of the timespan.    In this paper, we 
propose a  
co-evolutionary multi-task learning method that provides a synergy between 
multi-task learning and co-evolutionary algorithms to address dynamic time 
series prediction. The method features effective use of building blocks  of 
knowledge   inspired by dynamic programming and multi-task learning. It 
enables 
neural 
networks to   retain 
modularity  during training for  making a decision in situations even when 
certain inputs are missing.    The effectiveness of 
the method
is demonstrated using  one-step-ahead chaotic time series and tropical cyclone 
wind-intensity prediction.   
\end{abstract}


\end{frontmatter}

\section{Introduction}

Time series 
prediction  typically involves 
a pre-processing stage where the original time 
series is reconstructed into a state-space representation that is used as 
dataset for training models such as neural networks 
\cite{Mirikitani2010,ArdalaniFarsa2010, 
Teo2001,Gholipour2006,RutaEnsembles2007, 
Lin2009,Chandra2012times,ChandraTNNLS2015}. The reconstruction involves 
breaking 
the time series using overlapping  windows known as  
timespan   taken 
at regular intervals which defines the time lag   \cite{Takens1981}. The 
optimal values for  
 timespan and time lag are needed for effective 
prediction. These values   vary 
on the type of problem and require  costly 
computational evaluation for model selection; hence, some effort has  been 
made to address this issue. Multi-objective and competitive coevolution 
methods have been used to take advantage of different features from the 
timespan    during training 
\cite{NandC16FeatureS,ChandMO2014}. Moreover, neural network have been used  
for determining optimal timespan of selected  time series 
problems \cite{Maus2011Embed}.

In  time series  for natural 
disasters such as cyclones 
\cite{HeatPotential2012,Zjavka2016,Winds2014NeuralNet}, it is 
important to develop models that can make predictions dynamically, i.e. the 
model  has the ability to make a prediction as soon as any 
observation or data is 
available. The minimal value for the timespan can have huge impact 
for the case of cyclones, where data is only  available every 6 hours   
\cite{RatneelIJCNN2016}. A way to address such categories  of problems is 
to devise 
robust training algorithms and models that are capable of performing given 
different types of input or subtasks. 
We define \textit{ dynamic 
time 
series prediction} as a problem that requires dynamic  
prediction given a set of input features that vary in size. It 
has been highlighted in recent work \cite{RatneelIJCNN2016}  that recurrent 
neural networks trained with a predefined timespan   can only 
generalise well for the 
same timespan  which makes   dynamic time series 
prediction a challenging problem. 
Time series prediction problems can be generally characterised into three major 
types of problems 
that include   one-step   
\cite{Teo2001,ArdalaniFarsa2010,Chandra2012times}, multi-step-ahead  
\cite{taieb2015bias,chang2012reinforced,bone2002multi}, and multi-variate time 
series prediction \cite{Chakraborty1992,Wang2016MVTS,Zhang2016}. These problems 
at times may overlap with each other, for instance, a multi-step-ahead 
prediction can have a multi-variate component. Similarly, a one-step prediction 
can also have a multi-variate component, or a one-step ahead prediction can be 
used for multi-step prediction and vice-versa.  In this paper, we  
identify  a special class of problems that require dynamic prediction 
with the hope that the trained model can be useful for different instances of 
the problem.

Multi-task learning employs  shared representation knowledge for learning  
multiple 
instances from  the same problem  with the goal to develop   models with 
improved   performance in decision making
\cite{Caruana1997,evgeniou2005MT,MT-face-expZheng2016,MT-DeepNN2015}.
 We note that different values in the timespan can be 
used to generate several distinct datasets that have overlapping features which 
can  be  used to train modules for shared knowledge representation as needed 
for multi-task learning. Hence, it is important to ensure that modularity is 
retained   in such a way so that decision making can take place 
even when certain inputs are missing. Modular neural networks have been 
motivated from repeating 
structures in nature   and applied  for visual recognition 
tasks   \cite{Happel1994}.   
Neuroevolution   has been used to optimise  performance and connection 
costs  in modular neural networks \cite{Clune2012} which also has  the 
potential of  
learning 
new tasks without forgetting old ones \cite{Ellefsen2015}. The  features of 
modular learning provide motivation to be incorporated with multi-task learning 
for dynamic time series prediction.

  In dynamic programming, a  large problem is broken down into 
 sub-problems, from which at least one sub-problem is used as a  building 
block  for the  optimisation problem. Although dynamic programming has been 
primarily used for optimisation problems, it  
 has been briefly   explored for data driven  
learning  \cite{neuro-DP1997} \cite{datadrivenDP2015}. The notion of using 
sub-problems as building block in dynamic programming can be used in   
developing algorithms for  multi-task learning. Cooperative coevolution (CC) 
is a divide and conquer approach that divides a problem into subcomponents that 
are implemented as sub-populations \cite{Potter_Jong1994}. CC has been effective 
  for   learning difficult  problems using neural networks \cite{Potter2000}. 
Potter and De Jong  demonstrated that CC  provides more 
diverse solutions through the sub-populations when compared to conventional 
evolutionary algorithms \cite{Potter2000}. CC has been very effective for 
training recurrent neural networks for time series prediction    problems 
\cite{Chandra2012times,ChandraTNNLS2015}. 
 
  Although multi-task learning has mainly been used for machine learning 
problems, the concept of shared knowledge  representation has motivated other 
domains.  In the 
optimisation literature, 
multi-task evolutionary algorithms have been proposed  for   exploring and  
exploiting    
common knowledge 
between the tasks and enabling transfer of knowledge between them for 
optimisation \cite{Gupta2016TEC,YewSoon2016}. It was demonstrated that 
knowledge 
from related  tasks can help in speeding up the optimisation process and 
obtain 
better  quality solutions when compared to conventional (single-task 
optimisation)  
approaches.   Evolutionary multi-task learning has 
been used for efficiently training feedforward neural networks for   $n$-bit 
parity problem \cite{Chandra2017NPL}, where different subtasks were 
implemented as  different topologies  that obtained  improved training 
performance. In the literature, synergy 
of  dynamic programming, 
multi-task learning and neuroevolution has not been  explored.  Ensemble 
learning methods would be able to address 
dynamic time series  to an extent,  where an ensemble is defined by the 
 timespan of 
the time series. Howsoever, it would not have the feature of shared knowledge 
representation that is provided through multi-task learning. Moreover, there is 
a need for a  unified  model for dynamic times series problems  due to problems 
 that require dynamic prediction. 
 
In this paper, we propose a  
co-evolutionary multi-tasking method that provides a synergy between 
multi-task learning, dynamic programming and coevolutionary algorithms. The 
method   enables neural networks to 
be trained by featuring shared and modular knowledge representation  in order 
to make predictions given limited input features. 
This enables the learning process to employ modules of knowledge from  the 
related subtasks as building blocks of knowledge for a  unified  model. The 
proposed method 
is used for one-step-ahead chaotic time series problems using feedforward 
neural 
networks for  benchmark problems.    The  method 
is also  used for  
 tropical cyclone 
wind-intensity prediction and addresses the problem of minimal timespan where 
 dynamic prediction is required.  The paper extends results presented 
 in \cite{ChandraICONIP2017}.

The rest of the paper is organised as follows. Section 2 gives a background on 
multi-task learning,  
cooperative neuro-evolution, and time series prediction. Section 3 gives 
details of the co-evolutionary multi-task learning method for dynamic time 
series prediction. 
Section 4 presents the results with discussion and Section 5 presents the  
conclusions and directions for future research.

\section{Background and Related Work}

\subsection{Multi-task learning and applications}

 A number of approaches have been presented that 
considers multi-task learning  \cite{Caruana1997} for different types of 
problems that include 
supervised and unsupervised learning  
\cite{Ando2005,NIPS2008_3499,Chen2009ICML,ZhouNIPS2011}.    The 
major approach to address negative transfer  for multi-task learning has been 
through task grouping where 
knowledge transfer
is performed only within each group 
\cite{Zhang2010TML,Bakker2003}. Bakker \emph{et  al.} for instance, presented a 
 
Bayesian approach in which some of the model parameters were shared   and 
others 
 loosely connected through a joint prior distribution   learnt from the data 
\cite{Bakker2003}.  Zhang and Yeung presented a convex formulation for 
multi-task metric learning by modeling the task relationships in the form of a 
task covariance matrix \cite{Zhang2010TML}.  Moreover, Zhong \emph{et  al.} 
presented 
flexible 
multi-task learning framework to
identify latent grouping structures in order to restrict negative knowledge 
transfer \cite{Zhong2016}. Multi-task learning  has recently contributed to a 
number of successful 
real-world applications that gained better performance by exploiting shared 
knowledge for multi-task formulation.  Some of these applications  include 
 1)  multi-task 
approach for `` retweet'' prediction behaviour of individual  users  
\cite{Tang2015}, 2)
recognition of facial action units   \cite{Zhang2016}, 3) automated Human 
\textit{Epithelial Type 2 } (HEp-2) cell 
classification   \cite{Liu2016NC}, 4)   kin-relationship verification using 
visual 
features  \cite{Qin2016Kinship} and 5)  object tracking     
\cite{Zhang2015OTrack}.

\subsection{Cooperative Neuro-evolution}

 Neuro-evolution  employs evolutionary algorithms for training 
neural networks \cite{AngelineGNARL1994} which can be classified 
into   direct 
\cite{AngelineGNARL1994,SANE1997}  and indirect encoding strategies 
\cite{StanleyNEAT2002}. 
In direct  encoding,   every
connection and neuron is specified directly and explicitly in the genotype 
\cite{AngelineGNARL1994,SANE1997}.  In  indirect  
encoding, the genotype specifies rules or some other structure for generating 
the network
\cite{StanleyNEAT2002}. Performance of direct and indirect encodings varies for 
specific problems.  Indirect encodings seem very intuitive and have biological 
motivations, however, in 
several cases they have shown not to outperform direct encoding strategies 
\cite{Gomez_Schmidhuber2008,HeidrichMeisner2009}. 

Cooperative coevolution for training neural networks is known as cooperative 
neuroevolution \cite{Potter2000,MobNet2002,Chandra2012sep}. Although 
cooperative coevolution 
faced challenges in problem decomposition,  it showed promising features 
that included   modularity and diversity 
\cite{Potter2000}. Further challenges  have been in 
area of credit assignment  for subcomponents \cite{Potter2000,MobNet2002}, 
  problem decomposition, and 
adaptation due to problem of  separability that refer to grouping 
interacting or highly correlated variables
\cite{Chandra2012sep}. In  cooperative neuro-evolution, problem decomposition  
has a major effect in the training and generalisation performance. Although 
several decomposition strategies have been implemented that vary for different 
network architectures, the two established  decomposition methods   are 
those on the synapse 
  \cite{Gomez_Schmidhuber2008}  and neuron level \cite{chandra2010, 
Chandra2012sep,ChandraNSPRNN2011}. In synapse level, the  network is 
decomposed to its lowest 
level where each weight connection (synapse) forms a 
subcomponent \cite{Gomez_Schmidhuber2008,Lin2009}. In neuron level, 
the neurons in the network act as the reference 
point for each subcomponent \cite{ESP_Gomez1997,ChandraNSPRNN2011}. They 
have shown good 
performance in pattern classification problems \cite{GomezPhD2003,chandra2010, 
ChandraNSPRNN2011}. Synapse level  decomposition has shown good 
performance in control 
and time series prediction  problems 
\cite{Gomez_Schmidhuber2008,Lin2009,Chandra2012times}, however, they gave  poor 
performance for pattern classification problems  \cite{Chandra2012sep}. Chandra 
\emph{et  al.} 
applied neural and synapse level   decomposition  for chaotic time series 
problems using recurrent neural networks 
\cite{Chandra2012times}. Hence, it was established that synapse level encoding 
was 
more effective  for time series and control problems 
\cite{Gomez_Schmidhuber2008,Chandra2012times}.    Chandra later presented 
competition and collaboration with neuron and synapse decomposition strategies 
 during evolution which improved the performance further
\cite{ChandraTNNLS2015}.

\begin{algorithm}
\caption{Cooperative neuroevolution  }
\small
\label{ccframework}
\begin{algorithmic}

\STATE \textbf{Step 1:} Decompose the problem   (neuron  or synapse level 
decomposition)
 
\STATE \textbf{Step 2:} Initialise and cooperatively evaluate each 
sub-population\\

\FOR{  each \textit{cycle} until termination }
   \FOR{ each Sub-population}   
     \FOR{  $n$ Generations}
\STATE i) Select and create new offspring 

\STATE ii) Cooperatively evaluate the new offspring 

\STATE iii) Update sub-population\\

\ENDFOR
 
 \ENDFOR
 \ENDFOR
  
\end{algorithmic}
\end{algorithm}
 
\begin{figure*}[ht!]
\centering
\includegraphics[width=100mm ]{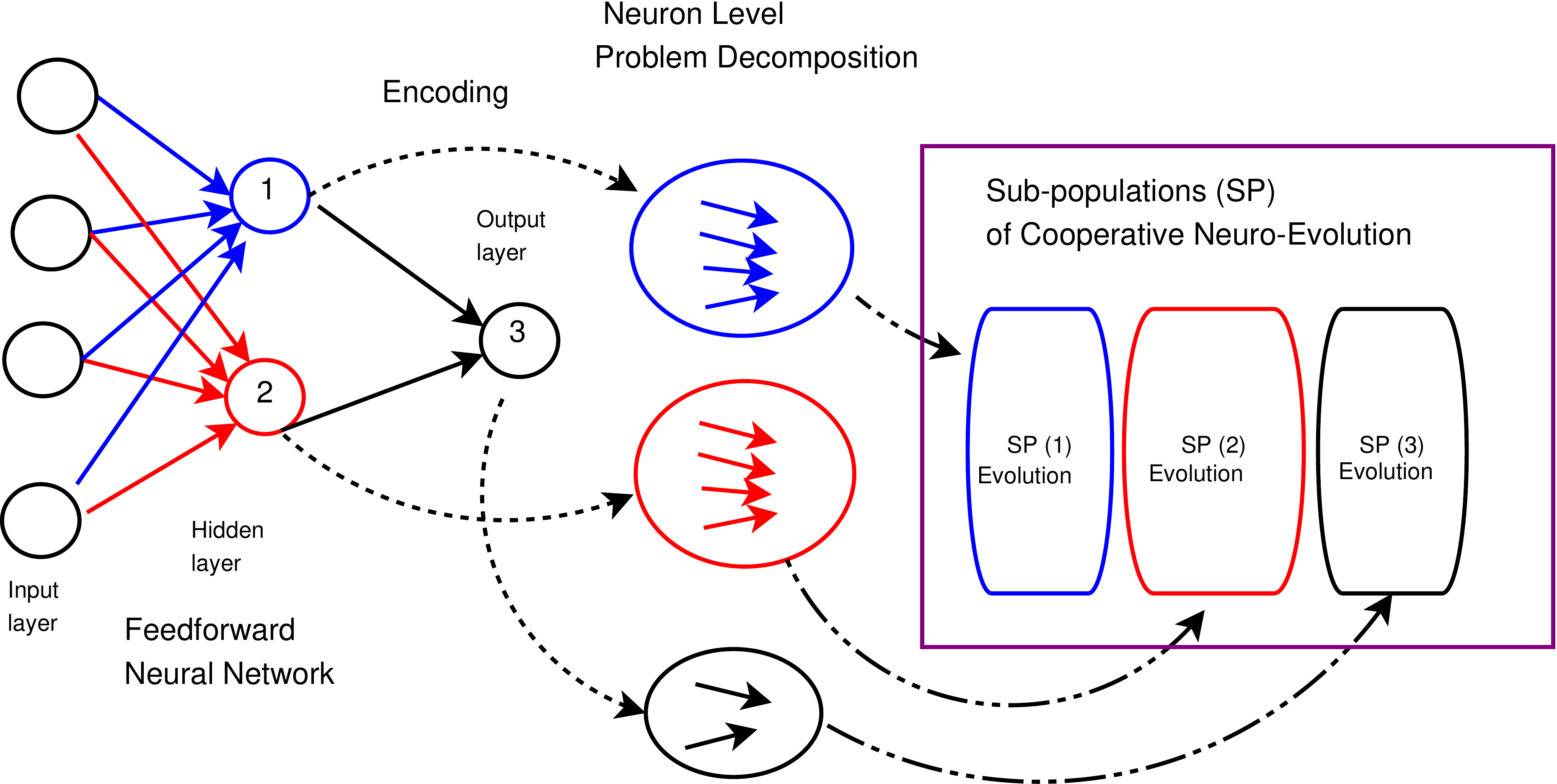}
\caption{Feedforward network with Neuron level   decomposition. Note that 4 
input neurons represent 
time series reconstruction with timespan of 4. }
\label{fnnNL}
\end{figure*}

In Algorithm \ref{ccframework}, the   network is decomposed according to 
the selected  decomposition method. Neuron level decomposition is shown 
in Figure \ref{fnnNL}. Once the decomposition is done, the subcomponents that 
are implemented as sub-populations are initialized and evolved in a round-robin 
fashion, typically for a fixed depth of search given by generations. 
The 
evaluation of the fitness of each individual for a particular sub-population is 
done cooperatively by concatenating the current individual with the fittest 
individuals from the rest of the sub-populations 
\cite{Potter2000}.   The concatenated individual 
is then  encoded  into the neural 
network where its fitness is evaluated and returned. Although   it is a 
representative fitness, the 
fitness of the entire network is   assigned to the particular 
individual of the sub-population. This is further illustrated  in Figure 
\ref{fnnNL}.

\subsection{Dynamic programming and reinforcement learning}
   Dynamic programming (also known as dynamic optimisation) is a 
optimisation strategy that considers breaking a large problem into sub-problems 
and using their solution as a building block to solve the 
bigger problem \cite{howard1966dynamic}.  By simply using  previously 
computed solution taken  from a sub-problem,   the paradigm  
improves the  computation time and  also  becomes efficient in memory or 
storage. 
 Although dynamic programming has typically been an approach for optimisation 
and sequential problems  \cite{held1962dynamic}, it has been well used in the 
areas of machine learning (such as spoken word recognition 
\cite{sakoe1978dynamic}),  and  computer vision (such as 
variational problems \cite{amini1990using}).   

   Reinforcement learning on the other hand,  considers agents 
that  take actions in an environment   to maximise the notion of cumulative 
reward \cite{sutton1992introduction}. Reinforcement learning has a wide range 
of multi-disciplinary applications  such as game theory, control theory, and
operations research \cite{kaelbling1996reinforcement}. In the operations 
research and control literature, reinforcement learning is called approximate 
dynamic programming \cite{bertsekas1995dynamic}, or neuro-dynamic programming 
\cite{bertsekas1995neuro}. They combine ideas from the fields of neural
networks, cognitive science and approximation theory. In machine learning, the 
environment is typically formulated as a Markov decision process (MDP) where 
the outcomes are partly random and  partly under the control of a decision 
maker. Opposed to classical dynamic programming,  reinforcement learning 
  does not assume knowledge of an exact mathematical model of the MDP. 
  Recent application  via deep learning 
considers learning of policies directly from high-dimensional sensory 
inputs  in the challenging domain of classic \textit{Atari 2600} games 
\cite{mnih2015human}. 

   Evolutionary algorithms have been proposed as  a method for  
reinforcement learning  \cite{moriarty1999evolutionary}. Reinforcement learning 
via evolutionary algorithms  have been implemented as neuroevolution 
\cite{richards1998evolving} for   neural networks  with application for 
playing the game of 'Go'. Reinforcement learning 
more recently has been implemented with co-evolutionary algorithms  in a 
classical  control problem  that considers balancing  double inverted poles
\cite{Gomez_Schmidhuber2008}. This further motivates the methodology presented 
in this paper that provides a synergy between, dynamic programming, 
reinforcement learning and neuroevolution for co-evolutionary multi-task 
learning.

 \subsection{Machine learning and optimisation}

Essentially, machine learning  algorithms have three components that include  
representation, evaluation and optimisation. Representation is done in the 
initial stage when the problem is defined and form formulated. Representation 
considers the type of problem (classification, regression or prediction) and 
the 
evaluation  metrics such as squared error loss and classification performance. 
Representation also considers initialisation of the parameters, such as the 
weights of the neural networks  and hyper parameters such as the learning rate. 
In the case of cooperative neuroevolution, the representation component would 
consider the encoding of the network weights into the subpopulations and 
initialising them for evolution.  Evaluation and optimisation are components 
that 
iterate over time until a certain condition is met.  Machine learning can also 
be seen as a data driven optimisation process. The learning procedure can be 
seen as 
solving a core optimisation problem that 
optimises the variables or parameters of the model with respect to the given 
loss 
function. Evolutionary algorithms are typically considered as optimisation 
methods and their synergy with neural networks into neuroevolution can be 
viewed 
as a learning procedure. In this paper,    learning in neural networks is 
implemented using   co-evolutionary algorithms that  features elements from 
multi-task learning and dynamic programming. Moreover, learning is also 
referred 
to evolution in the context of neuroevolution. Bennett and  Parrado-Hernández   
in an introductory note to a special issue of a journal mentioned  that 
optimisation problems lie at the heart of most machine learning approaches 
\cite{bennett2006interplay}. They highlighted the need for dealing with 
uncertainty, convex models, hyper-parameters,  and hybrid approaches of 
optimisation methods for learning. Furthermore, Guillory \emph{et al.} showed  
that  
online active learning algorithms can be viewed as stochastic gradient descent 
on non-convex objective functions \cite{guillory2009active}. 
  
 \subsection{Problems in time series prediction }

  Although a  number of methods have been used for one-step ahead  prediction,  
 neural networks have given  promising results   with different architectures 
\cite{ArdalaniFarsa2010,Chandra2012times} and algorithms that include 
gradient-based learning \cite{Koskela96,Mirikitani2010}, evolutionary 
algorithms \cite{Lin2009,Chandra2012times,ChandraTNNLS2015},  and hybrid 
learning methods 
\cite{Teo2001,Gholipour2006,ArdalaniFarsa2010}. These methods can also be used 
for multi-step ahead and multivariate time series prediction. Multi-step-ahead  
(MSA) prediction refers to  the forecasting or  prediction of 
 a 
sequence of 
future values from  observed trend in a time series 
\cite{sandya2013feature}.  It is challenging to develop models that 
produce low  
prediction error as the prediction horizon increases  
\cite{taieb2015bias,chang2012reinforced,bone2002multi}. MSA
prediction has been approached mostly with the \textit{recursive} and  
\textit{direct}  strategies. In the recursive strategy, the prediction from 
a one-step-ahead prediction model is used as input for future
prediction horizon
\cite{zhang2013iterated,ben2012recursive}. Although 
relatively new, a third strategy is a combination of these approaches   
\cite{zhang2013iterated,Grigorievskiy2014}.

Multi-variate time series prediction typically  involves the prediction of 
single or multiple values from multi-variate input that  are 
typically  interconnected through some event 
\cite{Chakraborty1992,Wang2016MVTS,Zhang2016}. Examples   of single value 
prediction  are the  prediction of flour  prices of time series obtained from   
different cities \cite{Chakraborty1992} and  traffic   time 
series \cite{Yin2016}. The goal in this case  is  to enhance 
the prediction performance from the  additional features in the input, 
although the problem can be solved in a univariate approach \cite{Yin2016}. In 
the case of prediction of multiple values, the model needs to predict future 
values of the different features, for example, prediction of 
latitude and longitude that defines the movement of cyclones 
\cite{ChandraDayalIJCNN2015}. A recent study has shown that     
multivariate 
prediction would perform  better than univariate for MSA  as the prediction 
horizon becomes larger, multi-variate information 
becomes more important 
\cite{Chayama2016}. Another area  of problems in  time series 
prediction 
 consist of  applications that have missing data.  Wu \emph{et  al.} approached 
the 
missing data problem in time series with non-linear filters and neural networks
\cite{Wu2014missingdate}.  In their method, a sequence of independent 
Bernoulli random variables  were used  to model random interruptions which was 
later  used to construct the state-space vector in pre-processing stage.

 Furthermore, novel approaches  that feature  a synergy of 
different methodologies  have recently been presented to address time series 
prediction. Extreme  value 
analysis considers the  extreme deviations 
from the median of probability distributions, which has been beneficial for 
time series prediction in the past \cite{smith1989extreme}. D'Urso \emph{et  
al.} 
explored the grouping 
  of time series with similar seasonal patterns using extreme value 
analysis with fuzzy clustering   with an application   daily 
sea-level time series in Australia \cite{Durso2017}.
Chouikhi \emph{et  al.} presented echo state networks for time series prediction 
where
particle swarm optimised was used   to optimise the 
untrained weights    that gave  enhancement  to  learning  \cite{Chouikhi2017}. 
Such approaches give motivations for developing a synergy of different 
methods in order to utilise their strengths and eliminate their weaknesses.   

\section{Co-evolutionary Multi-task Learning}

 \subsection{Preliminaries: time series reconstruction}


State-space reconstruction   considers the use of Taken's theorem 
which expresses that the
state-space vector  reproduces  important
characteristics of the original time series  \cite{Takens1981}. Given an 
observed time series $x(t)$, an embedded  state
space $ Y(t) = [(x(t),
x(t-T),..., x(t - (D-1)T)]$ can be generated, where, $T$ is the time delay, $D$ 
is
the timespan (also known as embedding dimension), $t=   (D-1)T,  DT, ..., N-1$, 
and $N$ is 
the 
length 
of the
original time series.    The optimal values for $D$ and $T$
must be chosen in order to efficiently apply Taken's theorem 
\cite{frazier2004}. Taken's proved that if the original attractor is of
dimension $d$, then $D = 2d+1$ will be sufficient to reconstruct the attractor
\cite{Takens1981}. In the case of using feedforward neural  networks, $D$ is 
the number of input neurons.

\subsection{Dynamic time series prediction}

Natural disasters such as torrential
rainfall, cyclones, tornadoes,  wave surges and 
droughts \cite{Calvo2000, YuanfeiBPNNCyclone2011, 
Winds2014NeuralNet,Zjavka2016}  require 
dynamic and robust  prediction models that  can make a decision as soon as  the 
event take place. Therefore,  if the model is trained over specific  
months for rainy seasons, the system should be able to make a 
robust prediction from the beginning of the rainy season. We define the \textit 
{event length} as the duration of an event which can be number of hours of a 
cyclone or number of days of drought or torrential  rain.

As noted earlier, in  a typical time series prediction problem, the 
original time series is reconstructed using Taken's theorem 
\cite{Takens1981,frazier2004}. In the 
case of cyclones, it is important to  measure  the performance of the  model 
when dynamic prediction is needed regarding track, wind or other 
characteristics of the cyclone \cite{RatneelIJCNN2016}. Dynamic prediction can  
provide early warnings to the community at risk.  For instance, data about 
tropical cyclone in the South Pacific is recorded at six hour intervals 
\cite{JTWC}. If the   timespan $D = 6$, the first prediction 
by the model at hand   would come after 36 hours which could have devastating 
effects.

 The  problem arises  when  the gap between each data point in the times series 
is a day or number of hours. The problem with the existing models such as 
neural 
networks used for cyclones  is the minimal
timespan $D$ needed to make a prediction. It has been  reported that 
 recurrent  neural 
networks trained with a given timespan (e.g. $D=5 $), cannot make 
robust
prediction for other timespan ( e.g. $D=7$ or $D=3$ )
\cite{RatneelIJCNN2016}. 
 Therefore, we introduce and define  the problem  of 
\textit{dynamic time 
series 
prediction}   that refers to the ability of a model to give a 
prediction given a set of timespan values rather than a single one.  This 
enables the model to make decision with minimum value of the timespan in cases 
when rest of features of data-points are not available.   A conventional 
one-step ahead time series 
prediction can be given by


\begin{align}
\label{eqn:eqlabel}
\begin{split}
 \textbf{x}  &=   x[t], x[t- 1], ..., x[t - D] \\
 xˆ[t+1] &= f(\textbf{x})
\end{split}
\end{align}

where $f(.)$ is a model such as a feedforward neural network and $D$ is a fixed 
value for the  timespan and  \textbf{x} refers to the input features. In the 
case 
of 
dynamic time series, rather than a single value, we consider a set of values   
for the timespan 

\begin{equation}
  \Omega_m =[D_1, D_2,...D_{M}]
\end{equation}

where $M$ is the number of subtasks, given  
$M \leq D $. Hence,  the input features for each subtask in  
dynamic time series prediction can be 
given by $\Psi_m$, where $m= 1, 2, ..., M $.

\begin{align}
\label{eqn:dts}
\begin{split}
     \Psi_m &=   x[t], x[t- 1], ..., x[t - 
\Omega_m]\\ 
\end{split}
\end{align} 

\subsection{Method}

In the proposed method, a co-evolutionary algorithm based on a dynamic 
programming 
strategy is proposed for multi-task learning. It features  problem 
decomposition in a  similar way as cooperative coevolution, however, the major 
difference lies in the way the solutions of the subcomponents are combined 
 to build to the final solution. Hence, the proposed co-evolutionary multi-task 
learning algorithm is inspired from the strategies used in dynamic programming 
where a subset of the solution is used as the main building block for the 
optimisation problem. In this case, the problem is  learning 
the weights of 
a cascaded neural network architecture where  the base problem is the   network 
module that is defined by lowest number of input features  and  hidden neurons.

\begin{figure*} 
\centering
\includegraphics[width=105mm]{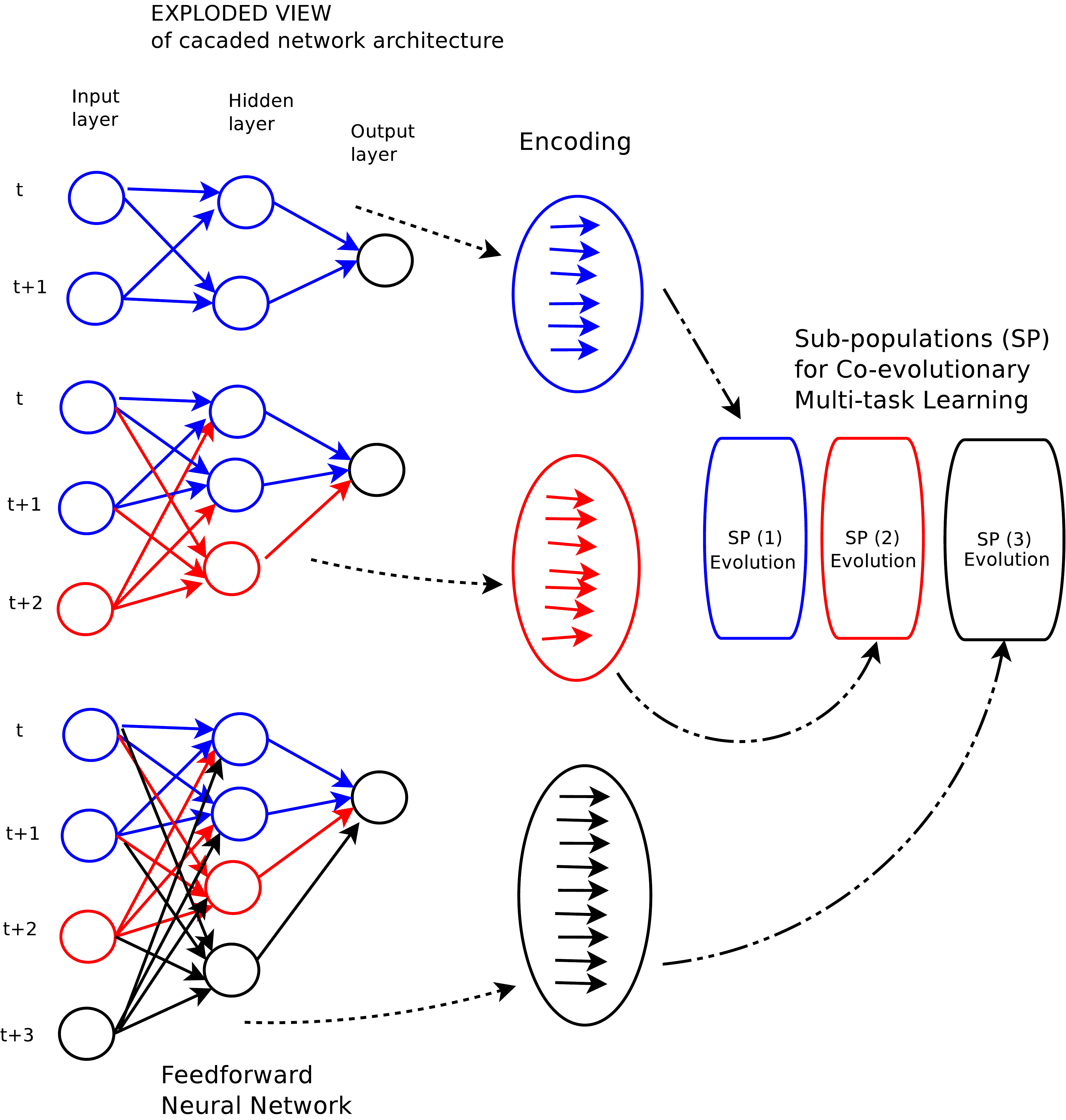}
\caption{Problem decomposition as subtasks in co-evolutionary multi-task 
learning. 
Note that the 
colours associated with the synapses in the network are linked to their 
encoding that are given as different subtasks. Subtask 1 employs a network 
topology 
with 2 hidden neurons while the rest of the subtasks add   extra input and 
hidden neurons. The exploded view shows that the different neural network 
topologies are assigned to the different subtasks, however, they are all part 
of 
the same network as shown in Figure \ref{transferMT} }
\label{encodeMT}
\end{figure*}

The weights in the base 
network are part of larger cascaded network modules that consist of additional 
hidden neurons and input features. This can be viewed as modules of knowledge 
that are combined for larger subtasks that use knowledge from smaller subtasks 
as 
building blocks. The cascaded network architecture can also be viewed as  an 
ensemble of neural networks that feature distinct topologies in terms of number 
input and hidden neurons as shown in Figure \ref{encodeMT}. Suppose that we 
refer to a module  in the cascaded ensemble, there are  $M$ modules 
with  input $i$, hidden $h$  layers  as shown.

\begin{align}
\label{eqn:eqlabel}
\begin{split}
  I  &= [i_1, i_2, ...,i_M] \\ 
 H &= [h_1, h_2, ...,h_M] \\ 
  O &= [1, 1, ..., 1]
\end{split}
\end{align}

\noindent where $I$, $H$, and $O$ contain the set of input, hidden and 
output layers.  Note that the 
approach considers fixed number of output neurons. Since 
we consider one-step-ahead time series problem, one neuron in output layer is 
used for all the respective modules. The input for each of the modules is given 
by the dynamic nature of the problem that considers different lengths of 
timespan that constructs an input vector  $\Psi_m$ for the given module as 
follows.

\begin{equation} 
  \Psi_m = x[t], x[t- 1], ..., x[t - I_m] 
\end{equation}
 
 Note that the input-hidden layer $\omega_m$ weights  and the 
hidden-output layer $\upsilon_m$ weights are combined for the respective module 
$m$. The base knowledge module is given as $\Phi_1  = [\omega_1, \upsilon_1]$. 
The 
\textit{subtask}  $\theta_m$  is defined as the  problem of training the 
respective  
knowledge 
modules $\Phi_m$  with given input $ \Psi_m$.   Note 
that 
Figure  \ref{transferMT}  explicitly shows the knowledge modules
of the network   for $\omega_2$ and $\omega_1$, 
respectively. The  knowledge module for each  subtask is constructed in a 
cascaded network architecture 
as follows.  
\begin{align}
\Phi_1 & = [\omega_1, \upsilon_1]; \;\; 
\theta_1=(\Phi_1)\nonumber\\
\Phi_2&=[ \omega_2,\upsilon_2 ];\;\;\theta_2 = [\theta_1,\Phi_2]\nonumber\\
&\vdots\nonumber\\
\Phi_M &=  [ \omega_M,\upsilon_M ];\;\;
\theta_M=[\theta_{M-1},\Phi_{M}]\nonumber \\
\end{align}
The   vector of knowledge modules considered for training or optimisation  is 
therefore
$\mathbf \Phi=(\Phi_1,\ldots,\Phi_M) $.

\begin{align}
\label{eqn:eqlabel}
\begin{split}
  y_1 &= f(\theta_1, \Psi_{1} ) \\
  y_2 &= f(\theta_2, \Psi_{2} ) \\
  &\vdots  \\
  y_M &= f(\theta_M, \Psi_{M} ) 
\end{split}
\end{align}

Given $T$ samples of data, the loss $L$ for sample $t$ can be calculated by 
root mean squared error.
\begin{equation} 
  L_t  = \sqrt {\frac{1}{M}  \sum_{m=1}^M 
\left(\hat{y}  -y_{m}\right)^2  }
\label{eq:error}
\end{equation}

\noindent where $\hat{y}$ is the observed time series and $y_{m}$ is the 
prediction given by   subtask $m$. The loss $E$ for the entire 
dataset (all subtasks) is given by 

\begin{equation}  
  E  = \frac{1}{T}  \sum_{t=1}^T L_t   
  \label{eq:loss}
\end{equation}

 The training of the cascaded network architecture involves  
decomposition  as subtasks through    co-evolutionary 
multi-task learning  (CMTL) algorithm.  The knowledge modules in subtasks 
denoted  by $ \Phi_m $ are implemented as  subcomponents    $ S_{1}, 
S_{2}, ..S_{M}$, where $M$ is number of 
subtasks.   The subcomponents are implemented as 
sub-populations consist of matrix of variables 
that feature  the weights and biases $S_m = \Pi_{i,j}$, where   $i$ refers 
to  weights and biases 
and $j$ refers to the  individuals. The individuals of the sub-populations 
are 
referred as genotype while  the 
corresponding network module are referred as  the phenotype. Unlike conventional 
transfer learning methods,  the transfer 
of knowledge here is done implicitly through the sub-populations in CMTL. The 
additional subtasks are implemented through the cascades that utilise knowledge 
from the base subtask.  The fitness of the cascade is evaluated by utilising 
the knowledge from the base subtask. This is done through CMTL where the best 
solution from the sub-population of the base subtask is concatenated with the 
current individual from the sub-population whose fitness needs to be evaluated. 
This is how transfer of knowledge is implicitly done through co-evolutionary 
multi-task learning.

\begin{algorithm}[!htb]
\smaller
 \KwData{Requires input  $\Psi_{m} $  taken from data 
$\Sigma$ for respective subtasks $\theta_m$.}
 \KwResult{  Prediction error $E$ for dynamic time series  } 
 
 
 \For{  each   module $m$   }{

 1. Assign fixed depth of search  $\beta$ that defines the 
number of 
generation of evolution for each subcomponent $S_{m}$;  (eg.  $\beta = 5$ )

   2. Define the weight space (input - hidden and hidden - output layer) for 
the different subtasks $\theta_m$ defined by the respective network module 
$\Phi_m  = [\omega_m, \upsilon_m]$.\;

3. Create $X_m$ vector of subtask solutions from $\Theta_m$.  Note 
that the size of  $X_m$ depends on number of  network module considered in the 
subtask.
 
  4. Initialise  genotype of  the sub-populations 
$S_{m_j}$  given by $j$ individuals in a range $[-\alpha, \alpha]$ drawn from 
a uniform distribution  where $\alpha$ defines the range.    
 }

 \While{each phase until termination }{

 \For{each subtask $m$ }{

 \For{each generation until $\beta$}{

\For{  each Individual $j$ in subpopulation $\Pi_{i,j}$   }{
** Stage 1:\\
 * Assign $B_m$ as  best  individual for subtask $m$ from subpopulation 
$\Pi_{i,j}$ \\
 \If{$m ==1 $}{
  *  Get the current  individual from  subpopulation; $V_1   = 
\Pi_{i,j}$ where     $i$ includes all    the  
variables (weights).  Assign the current individual as $X_1 = V_1$

 }
 \Else{
 *   Get the current   individual from  subpopulation; $V_m  = 
\Pi_{i,j}$. Append  the current subtask solution   with best solutions 
from previous subtasks, $X_{m} = [B_1,...,  B_{m-1}, V_m]$
where $B$ is the best individual from previous subtask and $V$ is the current 
individual that needs to be evaluated.

}
 ** Stage 2:\\
* Execute Algorithm \ref{alg:two}: This will encode the $X_{m}$ into knowledge
modules 
 $ \Phi=(\Phi_1,\ldots,\Phi_M) $ for subtasks $f(\theta_m, \Psi_{m} )$

*   Calculate subtask output $y_m = f(\theta_m, \Psi_{m})$  and evaluate the 
given
genotype  though the loss function  $ E  = 
\frac{1}{T}  \sum_{t=1}^T 
L_t   $, where $L_t$ is given in Equation \ref{eq:loss}. \\

 }

\For{  each Individual $j$ in $S_m$  }{
 * Create new offspring via evolutionary operators such 
selection,  crossover and mutation\\
 
}

* Update $S_m$
\;

  }
  
  }
 }
 
 * Test the obtained solution 
 
 \For{  each module $m$  }{

 1.  Load best solution $S_{best}$ from   $S_{m}$. \;
 
 2. Encode   into the respective
weight space for the subtask $\Phi_m  = [\omega_m, \upsilon_m]$.\;

3. Calculate loss $E$ based on the test dataset.

 }

 \caption{ Co-evolutionary multi-task learning }
 
\label{alg:one}
\end{algorithm}

Algorithm \ref{alg:one} gives details for CMTL which begins by 
initialising  the   the sub-populations   
defined by the subtasks   which feature  the knowledge modules
$\Phi_m$ and 
respective subtask input features $\Psi_{m}$. The sub-populations   are 
initialised 
with real 
values  $[-\alpha,\alpha]$ drawn from uniform distribution where 
$\alpha$ defines the range.   Once this has been done, the algorithm moves into 
the 
evolution phase where each subtask is evolved for a fixed number for 
generations 
defined by depth of search, $\beta$. The major  concern here is the 
way the phenotype is mapped into genotype where a group of weight matrices 
given by $\Phi_m  =[\omega_m, \upsilon_m]$ that makes up subtask $\theta_m$ are 
converted into vector $X_m$. Stage 1 in Algorithm \ref{alg:one} implements the  
use of  knowledge from 
previous subtasks through multi-task learning.  In the case if the  subtask is 
a base problem ($m 
==1$), then the subtask solution $X_{m}$ is utilised in a conventional matter 
where knowledge from other subtasks or modules are not required to reach a 
decision. Howsoever, given 
that the subtask is not a base problem,   the 
current  subtask individual $X_{m} $ is appended with best individuals
from the previous subtasks, therefore,   $X_{m} = [B_1,...,  B_{m-1}, V_m]$, 
where $B$ is the best individual from previous subtask and $V$ is the current 
individual that needs to be evaluated. This will encode  
$X_{m}$ into knowledge
modules $\mathbf \Phi=(\Phi_1,\ldots,\Phi_M) $ for the respective subtasks. The 
algorithm then calculates subtask network output or prediction 
$y_m = f(\theta_m, \Psi_{m})$  and evaluate  the  individual  
though the loss function  $ E  = \frac{1}{T}  \sum_{t=1}^T L_t   $, where $L_t$ 
is given in Equation \ref{eq:loss}. The subtask 
solution is passed to Algorithm \ref{alg:two} along with the network topology   
in order to decode the subtask 
solution 
into the 
respective weights of the network. This could be seen as the process of  
genotype to phenotype mapping.  This procedure is executed for every 
individual  in the sub-population.  This 
procedure 
is 
repeated for every sub-population   for different phases until the termination 
condition 
is 
satisfied. The termination condition can be either the maximum number of 
function evaluations or a minimum fitness value from the training or validation 
dataset. Figure \ref{encodeMT} shows an exploded view of the neural 
network topologies associated with the respective  subtasks, however, they are 
part of the cascaded  network architecture  later shown in Figure 
\ref{transferMT}. The  
way the subtask solution is decomposed and mapped into the network is given in  
Figure \ref{transferMT} and discussed
detail in the next section.

 The major difference in the implementation of CMTL when compared to  
conventional  cooperative neuroevolution  (Algorithm \ref{ccframework}) is by 
the way the problem is decomposed and the 
way the fitness for each individual is calculated. CMTL is motivated by dynamic 
programming approach where the best solution from previous sub-populations are 
used for cooperative fitness evaluation  for individuals in the current  
sub-population.  Howsoever, the current sub-population  does not use the 
best solution from future subpopulations. This way, the concept of utilising
knowledge from previous subtasks as building blocks is implemented . On the 
other hand, cooperative neuroevolution follows a divide and conquer approach 
where at any given  subpopulation, in order to evaluate the individuals,  the 
best individuals from the rest of the sub-populations are taken into account.

Finally, when the termination criteria has been met, the algorithm moves into 
the testing phase where the best solutions from all the different subtasks are 
saved and encoded into their respective network topologies. Once this is done, 
the respective subtask test data is loaded and the network makes prediction 
that is evaluated with loss  
$E$ given in Equation \ref{eq:loss}. Other measure of error can also be 
implemented. Hence, we have highlighted the association of 
every individual in the respective sub-populations  with  different subtasks in 
the multi-task learning  environment. There is transfer of knowledge in terms 
of weights from smaller to bigger networks as defined by the subtask with its 
data 
which is covered in detail in next section.  A Matlab   implementation of 
this algorithm with respective datasets used for the experiments  is 
given online 
\footnote{\href{https://github.com/rohitash-chandra/CMTL\_dynamictimeseries}{
https://github.com/rohitash-chandra/CMTL\_dynamictimeseries}}.

\subsection{Transfer of knowledge}

One challenging aspect of the Algorithm \ref{alg:one} is the transfer of 
knowledge represented by the  weights of the respective neural networks that is 
learnt  by the different subtasks in CMTL.  The cascading network architecture 
 increase in terms of the input and hidden neurons  with the subtasks. 
Algorithm \ref{alg:two}  implements transfer of knowledge given the changes of 
the 
architecture by  the different subtasks. The goal  is to transfer  weights 
that are mapped from different sub-populations defined by the subtasks.  The 
algorithm  is 
given 
input parameters which are 

\begin{enumerate} 
\item  The   reference to subtask $m$;  
 
\item The current subtask solution; if $m$ is base task,  $X_{m} = [V_{m}]$ 
else      $X_{m} = [B_{1},..., B_{m-1}, V_{m} ]$, otherwise.  
\item The topology of the respective cascaded neural module  for the 
different subtasks in terms of number of  input, hidden, and output neurons. 
\end{enumerate}
 
\begin{figure*} 
\centering
\includegraphics[width=150mm]{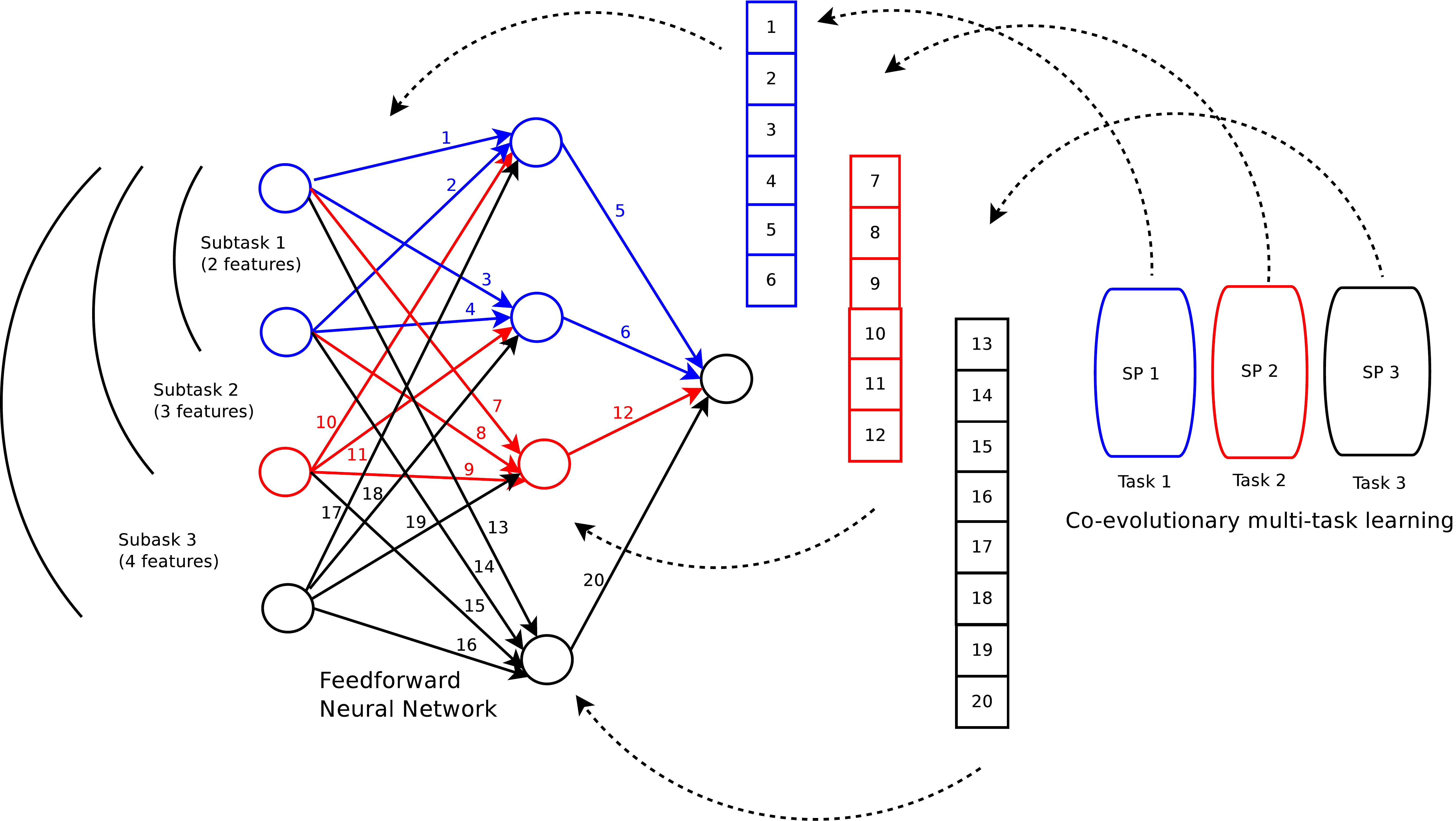}
\caption{Transfer of knowledge from subtasks encoding as sub-populations in 
co-evolutionary multi-task learning algorithm. This diagram shows transfer 
of 
knowledge from Subtask 1 and Subtask 2.   Once the  
knowledge is 
transferred into Subtask 3, the network  loads  Subtask 3 dataset  (4 
features 
in 
this example) for further evolution of the related sub-population   }
\label{transferMT}
\end{figure*}

We   describe the algorithm with reference to Figure \ref{transferMT} 
which shows a case, where the subtask $m = 3$ goes through the 
transfer where  $m = 1$ and $m = 2$ are used as building blocks of 
knowledge 
given  in the weights. Therefore, we use examples for the network topology 
as highlighted below. 

\begin{enumerate} 
 \item  $I_m$ is vector of number of input neurons for   the  
respective subtasks, eg. $ I = [2, 3, 4 ]$ ;
  \item  $H_m$ is vector of number of input neurons for  the respective 
subtasks, 
eg. $H_ = [2, 3, 4]$;
\item  $O_m$   is vector of  output neurons for   the respective subtasks, 
eg. 
$O = [1, 1, 1]$.  

\end{enumerate}

 The algorithm begins by assigning base case,  $b = 1$ which is 
 applied irrespective of the number of subtasks. In Step 1, the transfer 
of  \textit{Input-Hidden} layer weights  is   shown by weights (1-4)  in Figure 
\ref{transferMT}.  Step 2 
executes the transfer for   \textit{Hidden-Output} layer weights  as  
shown by weights (5-6)  in  Figure 
\ref{transferMT}. Note that Step 1 and 2 are applied for all the cases given by 
the number of subtasks. Once this is done, the algorithm terminates if  
$m = 
1$ or proceeds if $m >= 2$. Moving on, in Step 3, the 
case is more complex as we consider $m >= 2$.   Step 1 
and 
2 are executed before moving to Step 3 where $X$ contains the 
appended  solution sets from previous subtasks. In  Step 3,  $t$ in principle  
points to the 
beginning of the solution given by sub-population for $m = 2$. Here, the  
transfer for  \textit{Input-Hidden} layer weights  (7-9)  is executed 
for $m = 2$. Note that in this case, we begin with the weights with reference 
to the number of hidden neurons from previous subtask $j= H_{(m -1)} 
+ 1$, 
and move to the number of hidden neurons of  the current subtask $j= 
H_{(m)}$  in order to transfer the weights   to all the input 
neurons. This refers to \textit{ weights  (7-9)} in Figure \ref{transferMT}. 
Before 
reaching    transfer for $m = 3$, $m = 1$ and $m = 2$ 
transfer 
would have already taken place and hence the \textit{weights (13-16) } would be 
transferred as shown in the same figure. Moving on to Step 4,  we first 
consider the transfer for  \textit{Input-Hidden }
layer 
weights  for $m = 2$  through the transfer of weights  from beginning of 
previous subtask input,  $i=I_{(m-1)}+1$  to   current subtask 
input connected with all hidden neurons. This is   given by \textit{weights 
(10-11)} in 
Figure \ref{transferMT}. For the case of $m = 3$, this would refer to 
\textit{weights (17-19)} in the same figure.

  Finally, in Step 5, the algorithm executes the transfer for   Hidden-Output 
 layer weights based on the  hidden neurons from 
previous subtask. 
In 
case of $m = 2$,  this results in transferring \textit{weight (12)} and 
for 
$m = 
3$, the transfer is  \textit{weight (20)} in Figure \ref{transferMT}, 
respectively. Note that the algorithm can transfer any number of input and 
hidden neurons as 
the  number of  subtasks  increase.

\begin{algorithm}[!htb] 
 \smaller
 \textbf{Parameters: }
  Subtask $m$,   module subtask solution $X$  , Input $I$, Hidden $H$  and 
Output $O$ \\

$b = 1$; ( Base task)

Step 1

\For{each $j=1$   to $H_{b}$ } {
\For{each  $i=1$   to $I_{b}$ } {

    $W_{ij}  = X_t$  \; \\
    $t = t+1$ \; 
 }
 }  

 Step 2 

\For{each $k=1$  to $O_{b}$ } {
\For{each $j=1$   to $H_{b}$} {

    $W_{jk}  = X_t$  \; \\
    $t = t+1$ \; 
 }
 }

   
   \If{m $>=$ 2}{ 
   
  Step 3

\For{each  $j= H_{m -1} + 1$ to $H_{m}$ } {
\For{each  $i=1$   to $I_{m}$ } {

       $W_{ij}  = X_t$  \; \\
    $t = t+1$ \; 
 }
 }  
 
 Step 4
   
\For{each  $j=1$   to $H_{m}$ -1 } {
\For{each  $i=I_{m-1}+1$ to $I_{m}$} {

        $W_{ij}  = X_t$  \; \\
    $t = t+1$ \; 
 }
 } 
  
   Step 5 
    
  \For{each $k=1$  to $O_{m}$ } {
   \For{each $j=H_{m-1}+ 1$   to $H_{m}$} {

      $W_{j,k}  = X_t$  \; \\
     $t = t+1$ \; 
   }
   }

   }

 \caption{Transfer of knowledge from previous subtasks}
 
\label{alg:two}
\end{algorithm}

 The time complexity of CMTL   considers the time 
taken for transfer of solutions for different number of subtasks. We
note that the best case is when the subtask is the base subtask $(m = 1)$. 
Therefore, the worst case time complexity can be 
given by

\begin{align}
\label{eqn:time}
\begin{split} 
 T(m<=1) &= O(1) \\
 T(m) &= T(m-1) + T(m-2), ...,  + O(1)\\
 T(m) &= O(2^m)  
\end{split}
\end{align}

where $m$ refers to the subtasks.

\section{Simulation and Analysis}

This section presents an experimental study that compares the performance of 
CMTL with conventional neuroevolution methods.    The results are compared 
with  neuroevolution via evolutionary algorithm (EA)  and 
cooperative neuroevolution (CNE)  for benchmark time series prediction 
problems. Furthermore,  tropical 
cyclones from South Pacific and South Indian Ocean are considered  to address 
the minimal timespan issue  \cite{RatneelIJCNN2016} using dynamic 
time series prediction.

 \subsection{Benchmark Chaotic Time Series Problems  } 

In the benchmark chaotic time series problems, the Mackey-Glass, Lorenz, Henon 
 and Rossler are the four synthetic time series problems. The experiments use 
the chaotic time series with length of 
1000 generated by the respective chaotic attractor.  The first 500 samples are 
used for training and the remaining  for testing.

 In 
all cases, the phase space of the original time series 
is reconstructed  with the timespan for 3 datasets for the 
respective subtasks with the set of timespan $ \Omega = [3, 5, 7]$ and time 
lag $T = 2$. All the synthetic and  
real-world  time series were 
scaled in the range [0,1]. Further details of each of the time series problem 
is given as follows. 

The \textit{ Mackey-Glass time series}    has been used in literature as a 
benchmark problem due to its chaotic nature \cite{Mackey1977}. The 
\textit{Lorenz time series } was introduced by Edward Lorenz  who has
extensively contributed to the establishment of Chaos theory 
\cite{lorenz1963}. The \textit{Henon time series } is generated with a Henon 
map which is a discrete-time dynamical system  that exhibit chaotic behaviour 
\cite{Henon1976} and the \textit{Rossler time series } is generated using the 
attractor for the Rossler system, a system of three non-linear ordinary 
differential equations as given in \cite{rossler}. The real-world problem  are 
the Sunspot, ACI 
finance  and Laser time series. The \textit{Sunspot time series} is a good 
indication of the solar activities for solar cycles which impacts Earth's 
climate, weather patterns, satellite and space missions \cite{Sunspot2001}.   
The Sunspot time series from November 1834 to June 2001 is selected which 
consists of 2000 points.     The \textit{ACI financial time series } is obtained 
from the \textit{ACI
Worldwide Inc.}  which is one of the companies
listed on the NASDAQ stock exchange.  The data set contains
closing stock prices from December 2006 to February 2010,
which is equivalent to approximately 800 data points.  The closing
stock prices were normalised between 0 and 1. The data set
features  the recession that hit the U.S. market in 2008   
\cite{timeDataSet}. The \textit{Laser time series } is   measured in a 
physics laboratory experiment that were used in the \textit{Santa Fe 
Competition} \cite{lazerdata}. All the real world time series used the first 
50 percent samples for training and remaining for testing.

 \subsection{Cyclone time series }

 The Southern Hemisphere tropical cyclone best-track data from 
Joint Typhoon 
Warning Centre  recorded every 6-hours is  used as the main source of data 
\cite{JTWC}.  We consider only the austral summer tropical 
cyclone season 
(November to April) from 1985 to 2013 data  in the current study as data prior 
to the satellite era is  not reliable due to inconsistencies and missing 
values. The original data of tropical cyclone wind intensity in the South 
Pacific 
  was divided into training  and testing set as follows:  
 
 \begin{itemize}
 \item Training Set: Cyclones from 1985 - 2005 (219 Cyclones with 6837 data 
points) 
 \item Testing  Set: Cyclones from 2006 - 2013  (71 Cyclones with 2600 data 
points)
  \end{itemize}
  
In the case for South Indian Ocean, the details are  as follows: 
 
 \begin{itemize}
 \item Training Set: Cyclones from 1985 - 2001 (  285 Cyclones with 9365 data 
points) 
 \item Testing  Set: Cyclones from 2002 - 2013  (  190 Cyclones with 8295 data 
points )
  \end{itemize}
  
  Although the cyclones are separate events,   we choose to  combine 
all the 
cyclone data  in a consecutive order as given by their 
date of occurrence. The time series 
is reconstructed  with the set of  timespan,   $ \Omega = [3, 5, 7]$, and 
time 
lag   $T = 2$.

 \subsection{Experimental Design}

  We note that multi-task learning approach used 
for dynamic time series can be formulated as a series of independent single 
task learning approaches. Hence, for comparison of CMTL,  we provide 
experimentation and results with conventional  neuroevolution methods that 
can be considered as single-task learning   (CNE 
and EA). In the case of CNE, neuron level problem decomposition 
is applied for training   
feedforward networks \cite{Chandra2012sep}.   We employ 
covariance matrix adaptation evolution 
strategies (CMAES) 
\cite{HansenMK03} as the evolutionary algorithm in sub-populations of CMTL, 
CNE and the population of EA.  The  training and generalisation 
 performances are reported for each case given by the different subtasks in the 
respective time series problems.

The respective neural  networks used both  sigmoid units in the hidden and 
output layer 
for all the different problems.   The loss function given in Equation 
\ref{eq:loss}  is used as the main performance measure.  Each neural 
network architecture was tested with different numbers of hidden neurons.

We employ fixed $depth = 5$  generation in the sub-populations of CMTL as the 
\textit{depth
of search} as it has given  optimal performance in trial runs. CNE also employs 
the same value.   Note that all the sub-populations   evolve for the same depth 
of search. The population size of CMAES in the respective methods is given 
by $P = 4+ floor*(3*log(W)) $, where $W$ is the total number of weights and 
biases   for the 
entire neural network  that includes all the subtasks (CMTL). In the case of EA 
and CNE,  $W$ is total number of weights and biases for the given network 
architecture.

The 
termination condition is fixed at 30 000 function evaluations for each subtask, 
hence, CMTL employs 120 000 function evaluations while  conventional methods
use 30 000 for each of  the respective subtasks for all the problems. Note that 
since there is a fixed training time, there was no validation set used to stop 
training. The choice of the parameters such as the appropriate 
population size, termination condition has been determined  in trial 
experiments. The experiments are well aligned with  experimental setting  
 from
previous 
work  \cite{Chandra2012times,chandra2015competition}.

\subsection{Results for Benchmark Problems}
 
\begin{figure}[ht!]
\centering
\includegraphics[width=55mm, angle= 270]{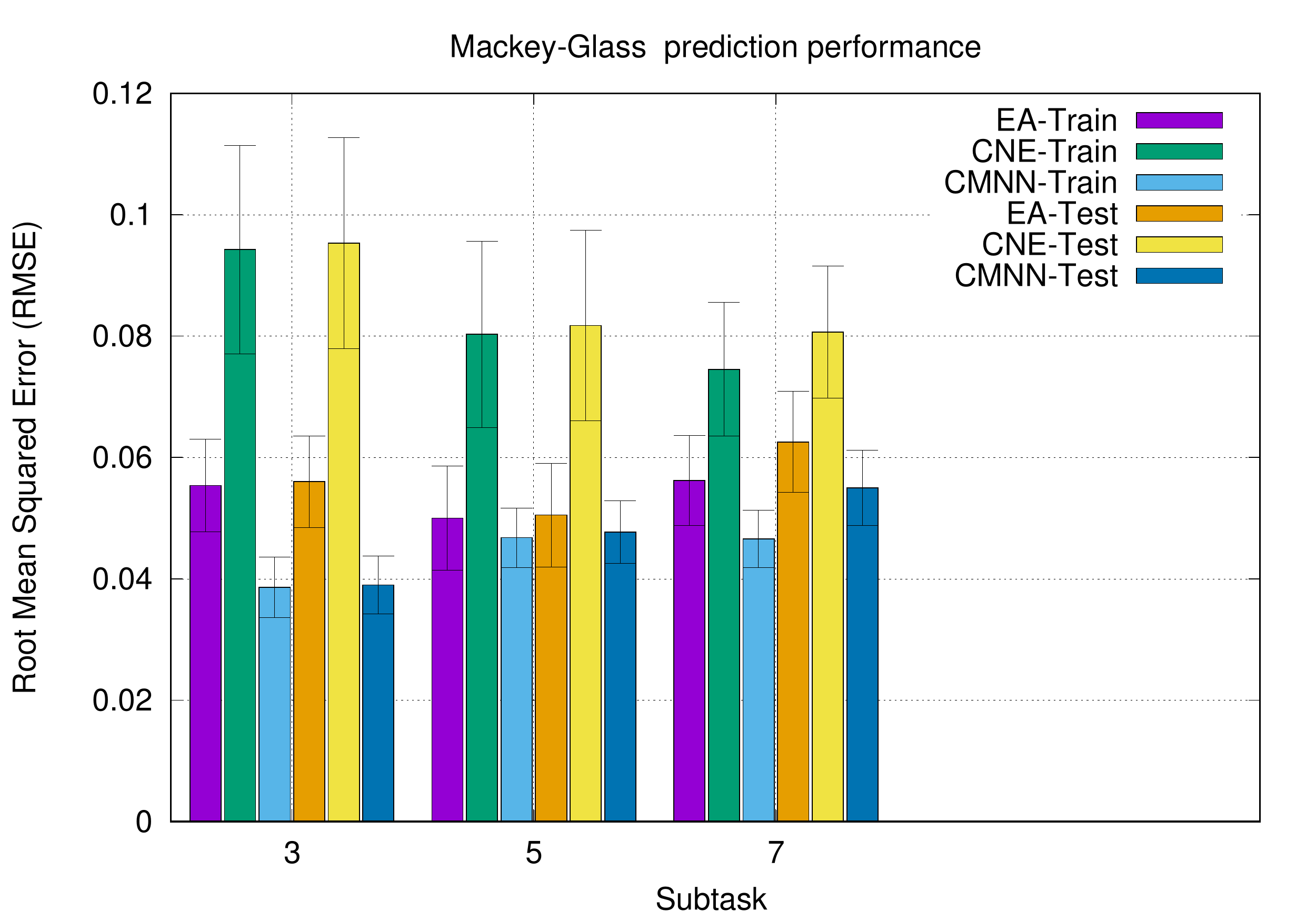}
\caption{Performance given by  EA, CNE, CMTL for Mackey-Glass time series}
\label{fig:mackey}
\end{figure}

\begin{figure}[ht!]
\centering
\includegraphics[width=55mm, angle= 270]{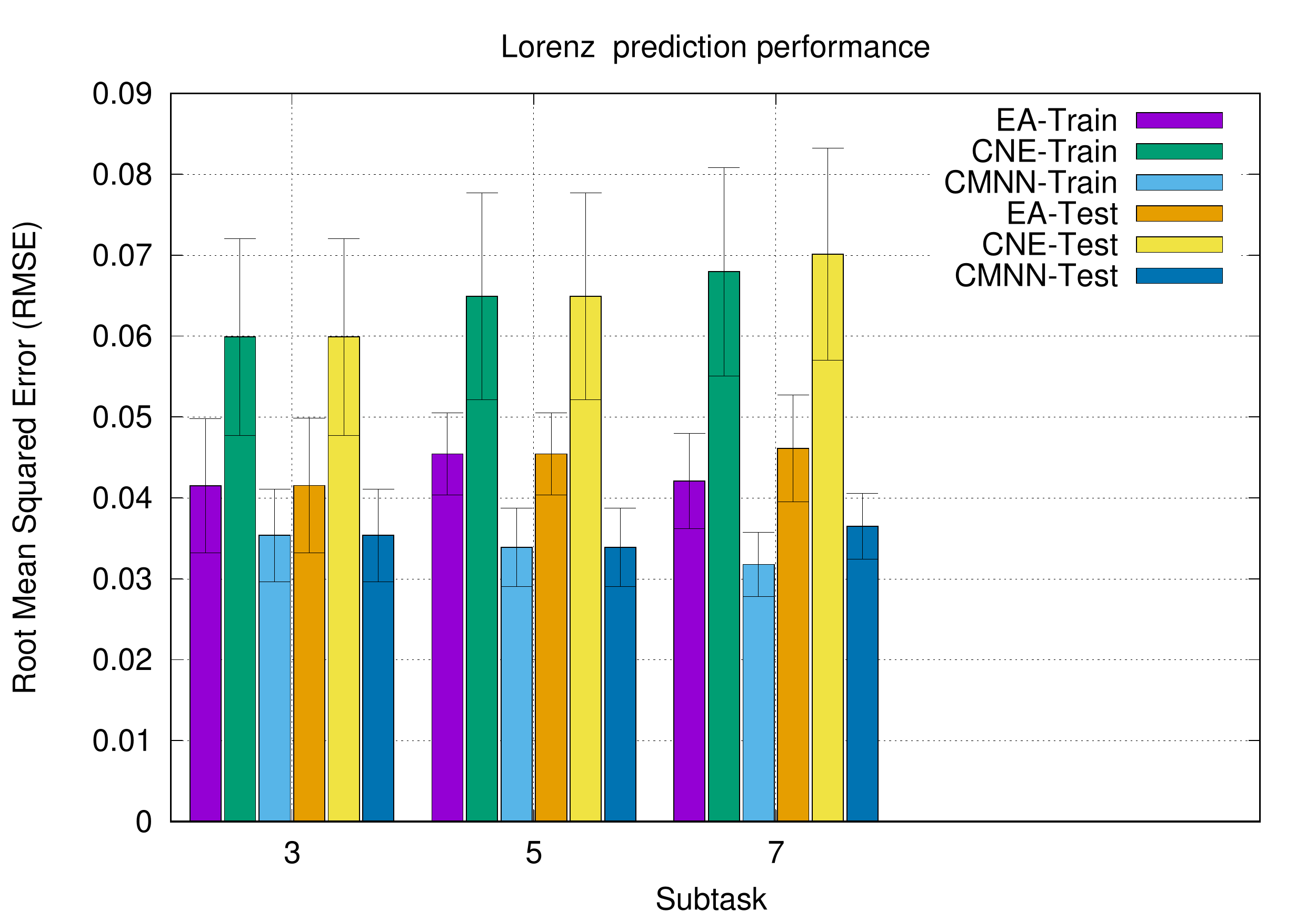}
\caption{Performance given by  EA, CNE, CMTL for Lorenz time series}
\label{fig:lorenz}
\end{figure}

\begin{figure}[ht!]
\centering
\includegraphics[width=55mm, angle= 270]{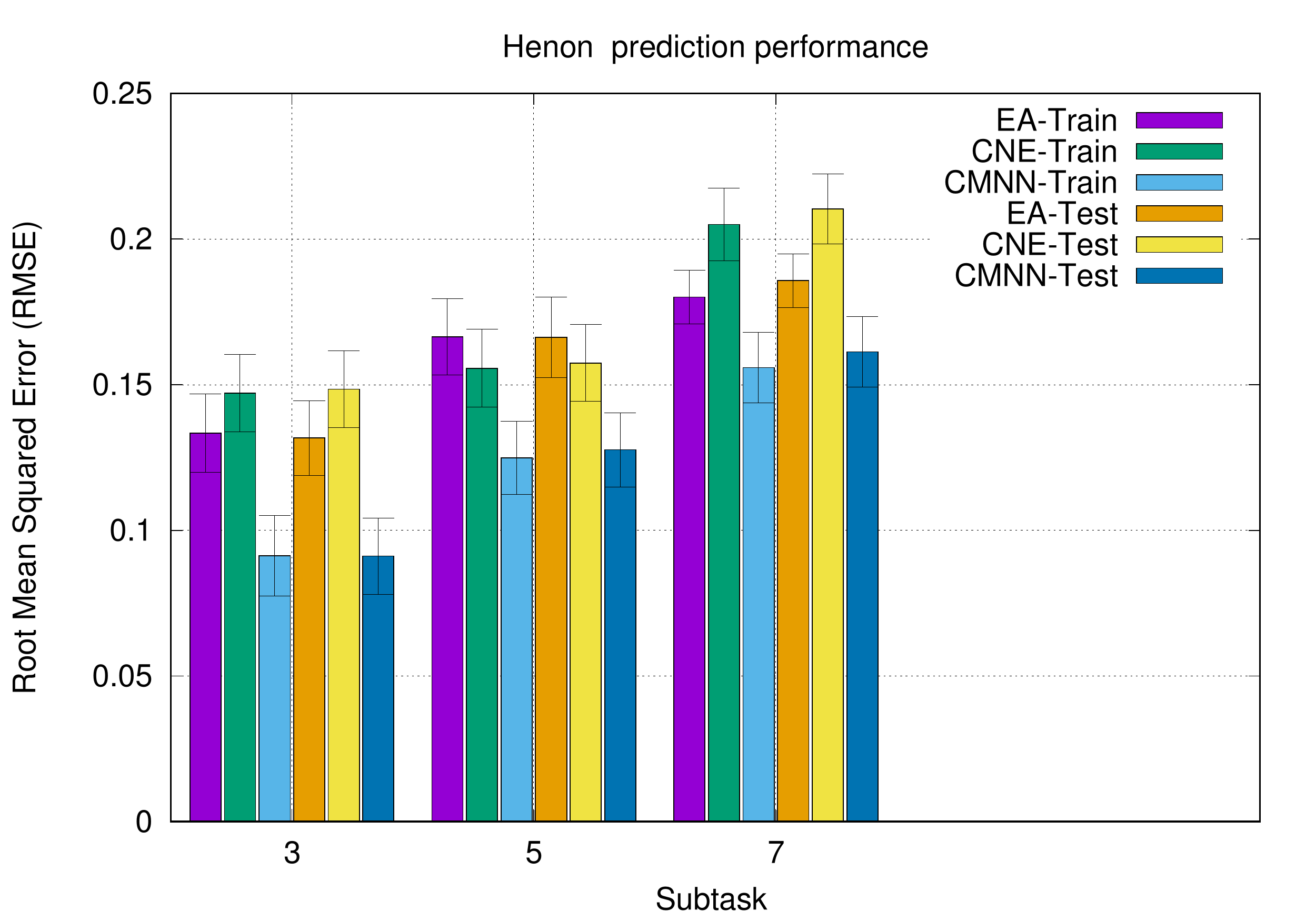}
\caption{Performance given by  EA, CNE, CMTL for Henon time series}
\label{fig:henon}
\end{figure}

\begin{figure}[ht!]
\centering
\includegraphics[width=55mm, angle= 270]{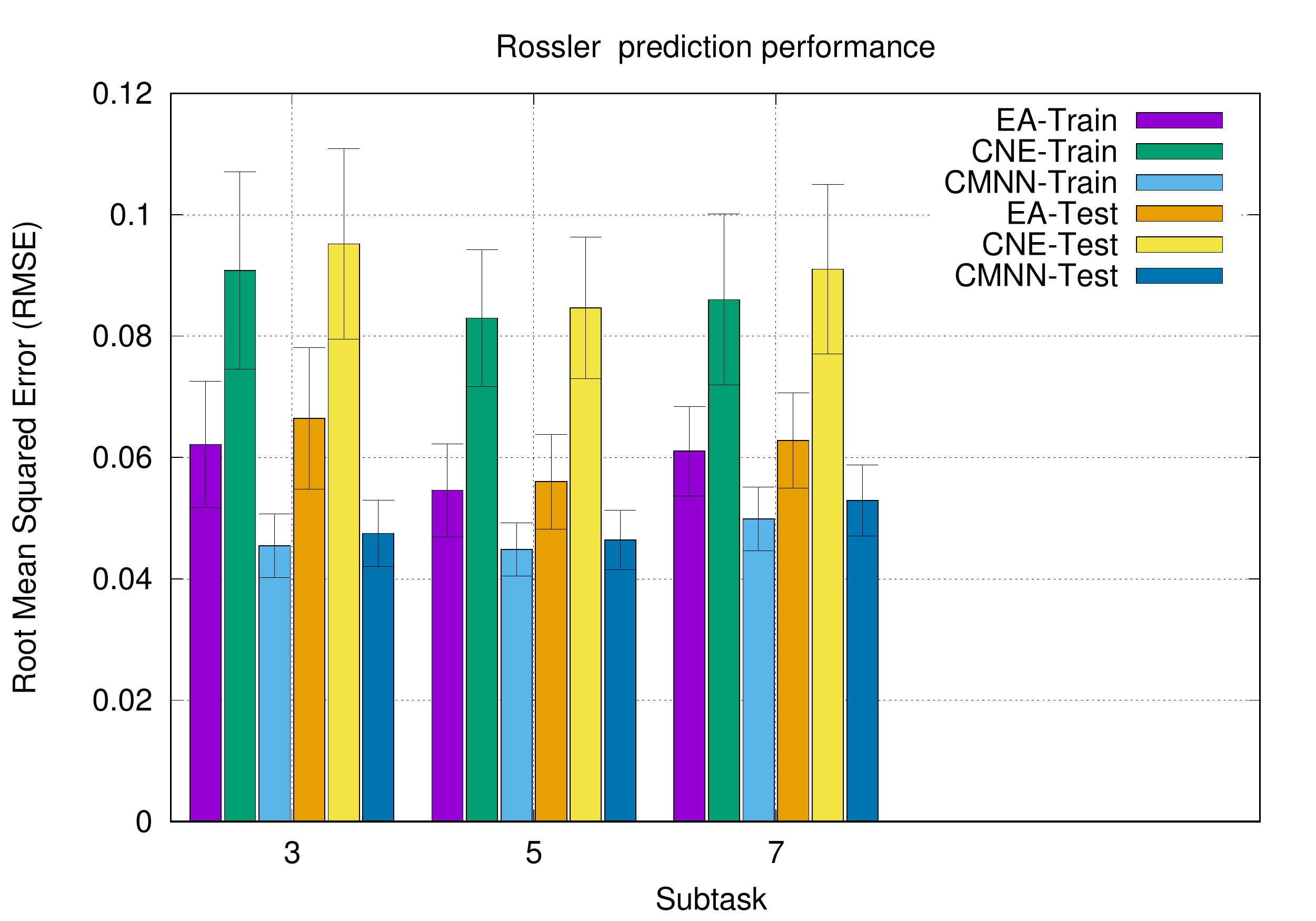}
\caption{Performance given by  EA, CNE, CMTL for Rossler time series}
\label{fig:rossler}
\end{figure}

\begin{figure}[ht!]
\centering
\includegraphics[width=55mm, angle= 270]{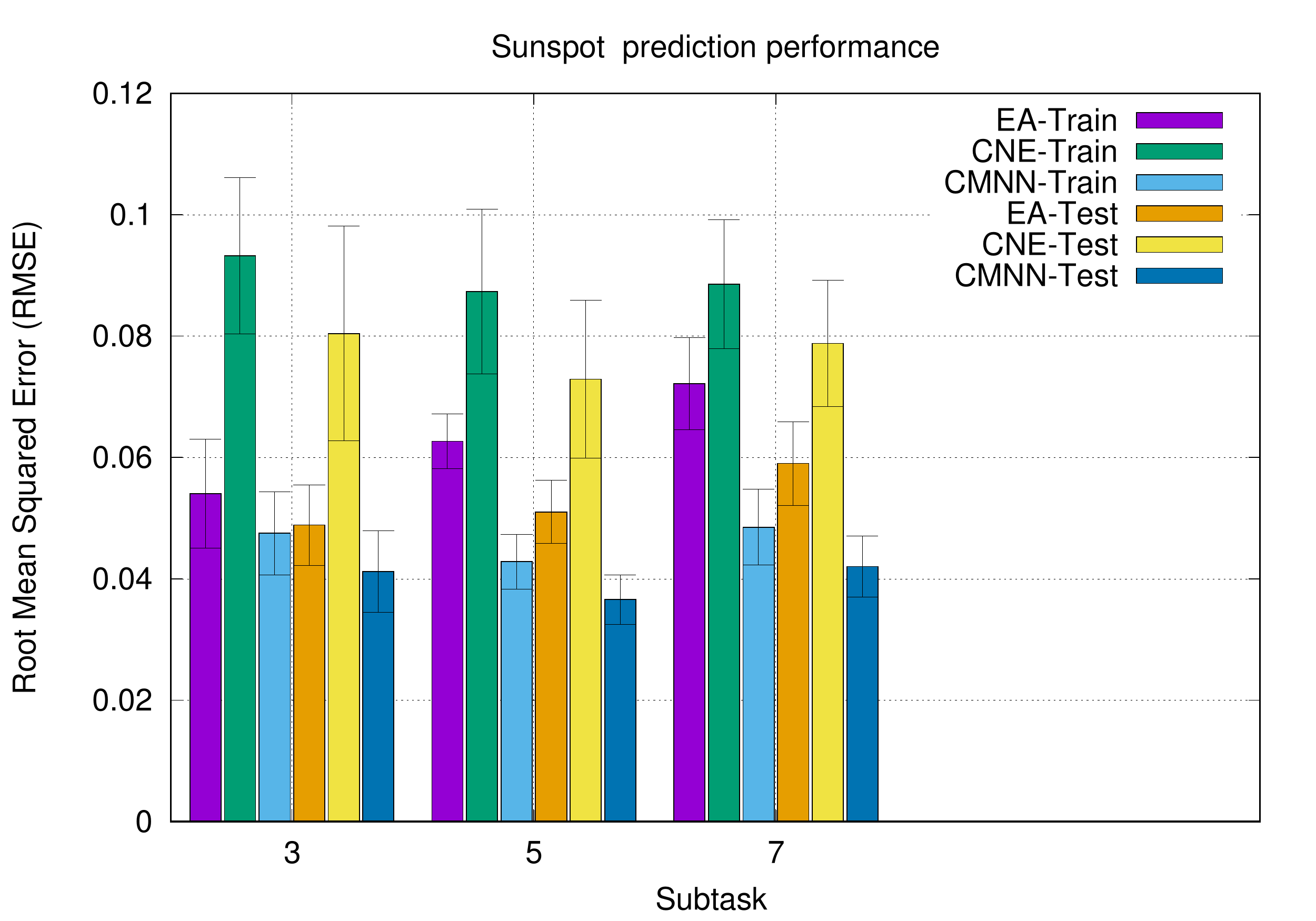}
\caption{Performance given by  EA, CNE, CMTL for Sunspot time series}
\label{fig:sunspot}
\end{figure}

\begin{figure}[ht!]
\centering
\includegraphics[width=55mm, angle= 270]{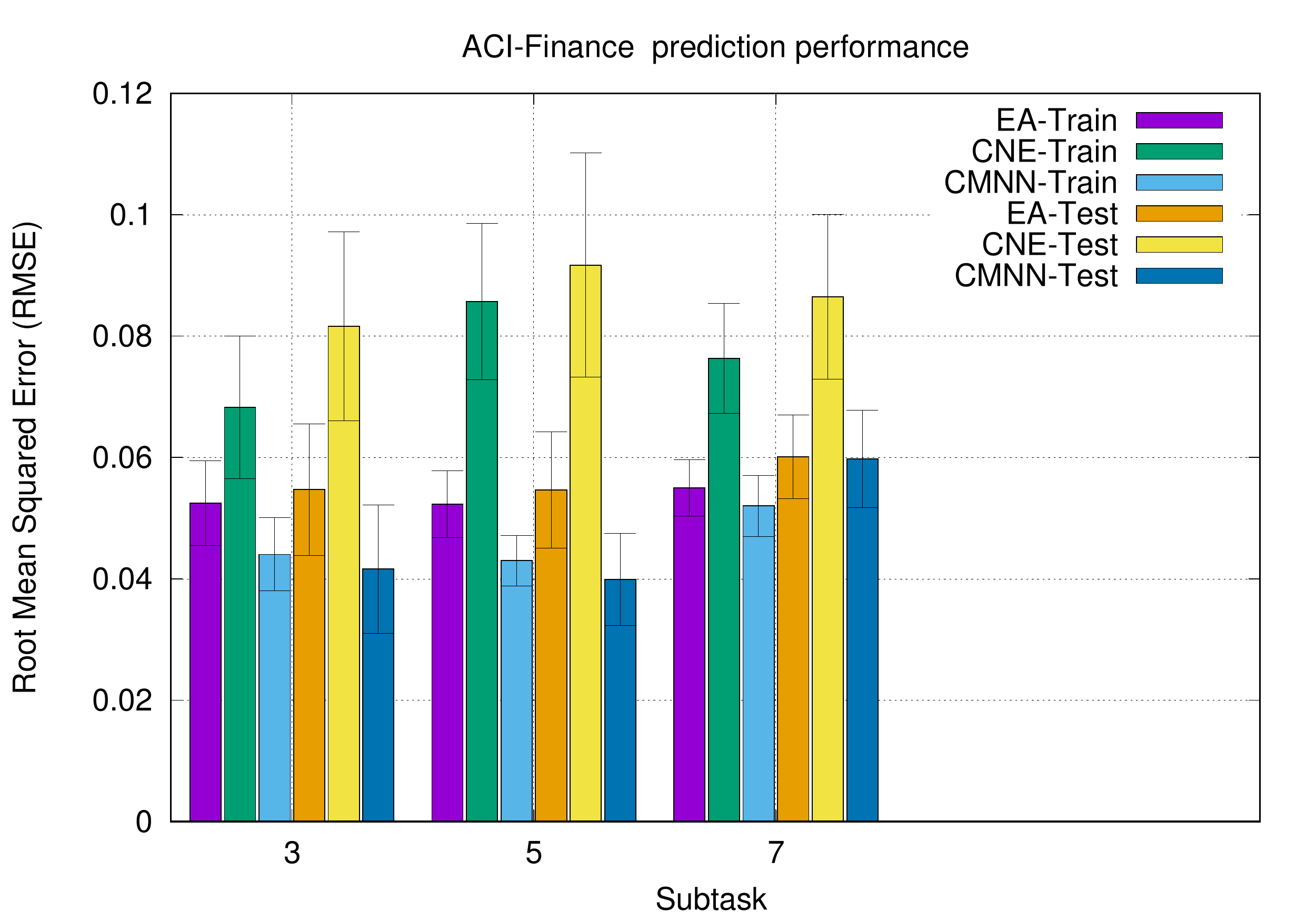}
\caption{Performance given by  EA, CNE, CMTL for ACI-finance time series}
\label{fig:aci}
\end{figure}

\begin{figure}[ht!]
\centering
\includegraphics[width=55mm, angle= 270]{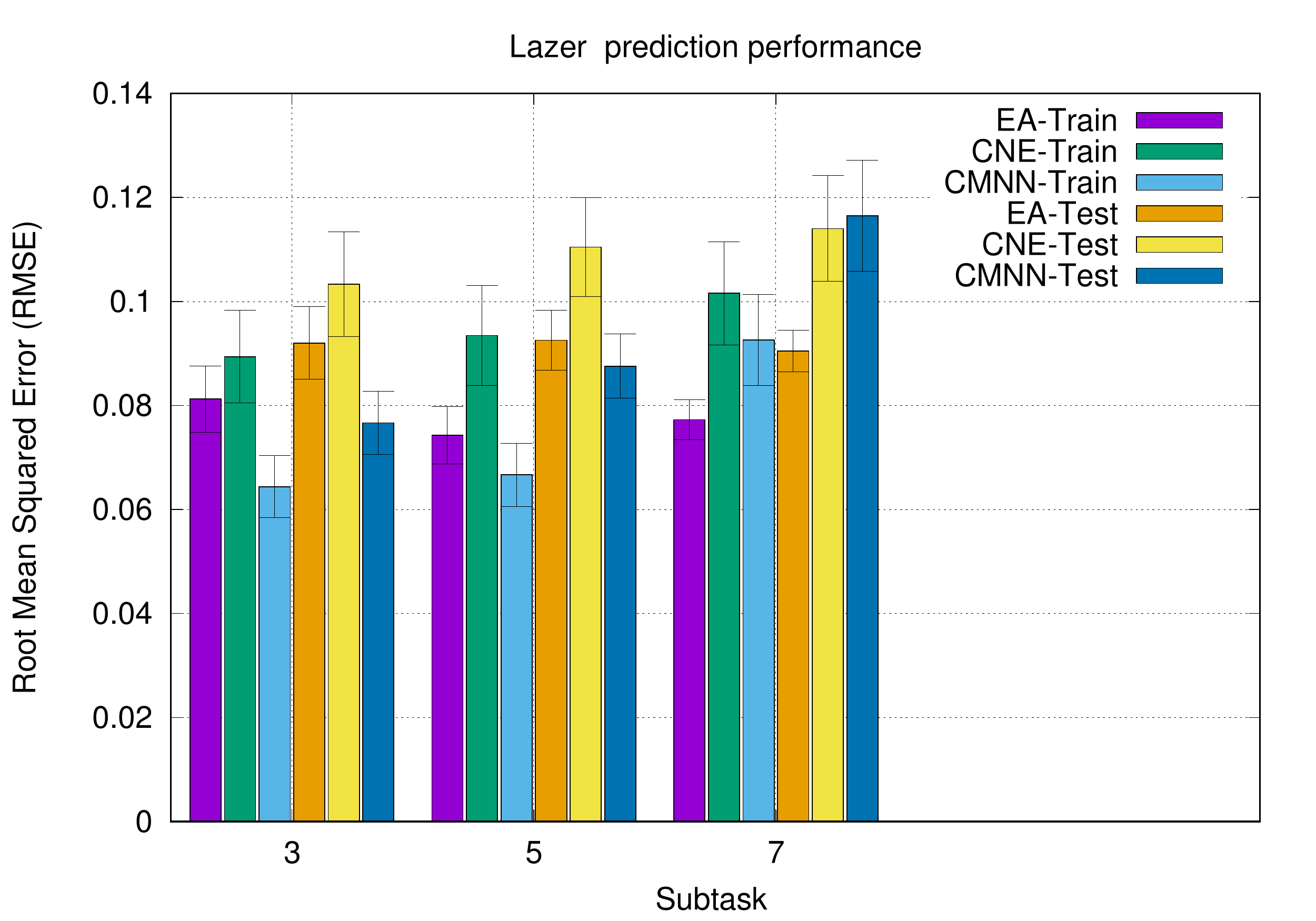}
\caption{Performance given by  EA, CNE, CMTL for Laser time series}
\label{fig:lazer}
\end{figure}

   The results for the 7 benchmark chaotic time series problems are given in 
Figure 
\ref{fig:mackey} to \ref{fig:lazer} which highlight the training  and 
generalisation performance.  We limit our discussion to  
the 
generalisation performance, although the training performance is also 
shown. Figure \ref{fig:mackey} shows that CMTL generalisation  performance is 
better 
than  EA and CNE, while  CMTL and EA outperform   CNE in all the subtasks 
denoted by the timespan. The same trend 
is shown in general  for Lorenz and Henon time series as shown in Figure 
\ref{fig:lorenz} and Figure \ref{fig:henon}, respectively. There is one 
exception ($D=5$), for the Henon problem  where CME gives better performance 
than 
EA, however, worse than CMTL. Figure \ref{fig:rossler} shows the results for 
the Rossler time series which follows a similar trend when compared to the 
previous problems. Hence, in general,  CMTL 
generalisation performance is the best when compared to the conventional
methods (CNE and EA) for the 4 synthetic
time series problems which have little or no   noise present. 

In the case of real-world problems, Figure \ref{fig:sunspot} for 
the Sunspot problem shows that CMTL provides the best generalisation 
performance when compared to EA and CNE for all the cases. The same 
is given for first two timespan cases for ACI-finance problem as shown in 
Figure \ref{fig:aci}, except for one case  ($D=7$), where EA  and CMTL gives 
the same performance. In the case of the Laser time series in Figure 
\ref{fig:lazer}, which is known as one of the most chaotic time series 
problems, CMTL outperforms CNE and EA, except for one case, $D=7$. Therefore, 
at this 
stage, 
we can conclude that CMTL gives the best performance for most of the cases in 
the real-world time series problems. Table \ref{tab:MeanRes} shows the mean of 
RMSE and confidence interval  across 
the 3 timespan. We find that the CMTL performs better than EA 
and CNE for almost  all the problems. The  Laser problem is  the only 
exception where the EA is slightly better than CMTL.

\begin{table*}[htb!]
\centering
 \small
 \caption{Combined mean prediction performance  for  3 subtasks}
\label{tab:MeanRes}

\begin{tabular}{llll}
 \hline
Problem& EA  &CNE & CMTL \\
 \hline
 \hline
Mackey-Glass  & 0.0564		$\pm $0.0081 &	0.0859		$\pm $0.0147 & 
0.0472		$\pm $0.0054\\

Lorenz&    0.0444		$\pm $0.0067 &	0.0650	$\pm $	0.0127 &	
0.0353	$\pm $	0.0049\\

Henon & 0.1612		$\pm $ 0.0120 &	0.1721	$\pm $	0.0128 &	
 0.1267	0.0127 \\

Rossler    &  0.0617$\pm $	0.0091 &	0.0903$\pm $	0.0138 &	
0.0489	$\pm $0.0054 \\ 

 \hline
Sunspot&    0.0529 $\pm $	0.0062 &	0.0773 	$\pm $0.0137 &	
0.0399 	$\pm $ 0.0052 \\

Lazer & 0.0917	$\pm $0.0056	& 0.1093	$\pm $0.0099 &0.0936	$\pm 
$0.0077 \\

ACI-finance    &0.0565	$\pm $0.0091 &	0.0866	$\pm $0.0159&	0.0471	$\pm 
$0.0087\\

 \hline
 \hline

\end{tabular}
\end{table*}

%

\subsection{Results for Tropical Cyclones}
\begin{figure}[ht!]
\centering
\includegraphics[width=55mm, angle= 
270]{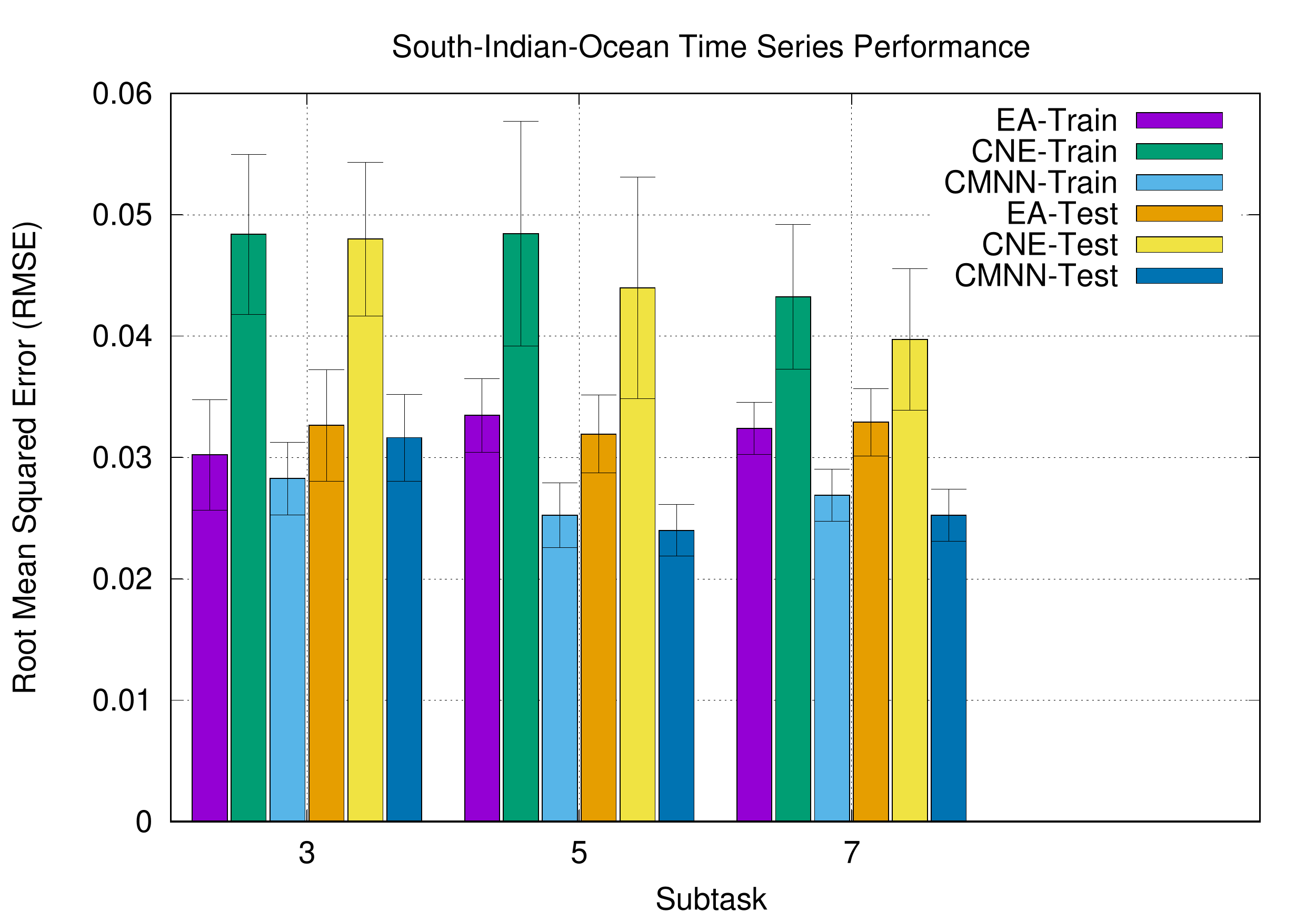}
\caption{Performance given by  EA, CNE, CMTL for South Indian ocean}
\label{fig:io}
\end{figure}

\begin{figure}[ht!]
\centering
\includegraphics[width=55mm, angle= 
270]{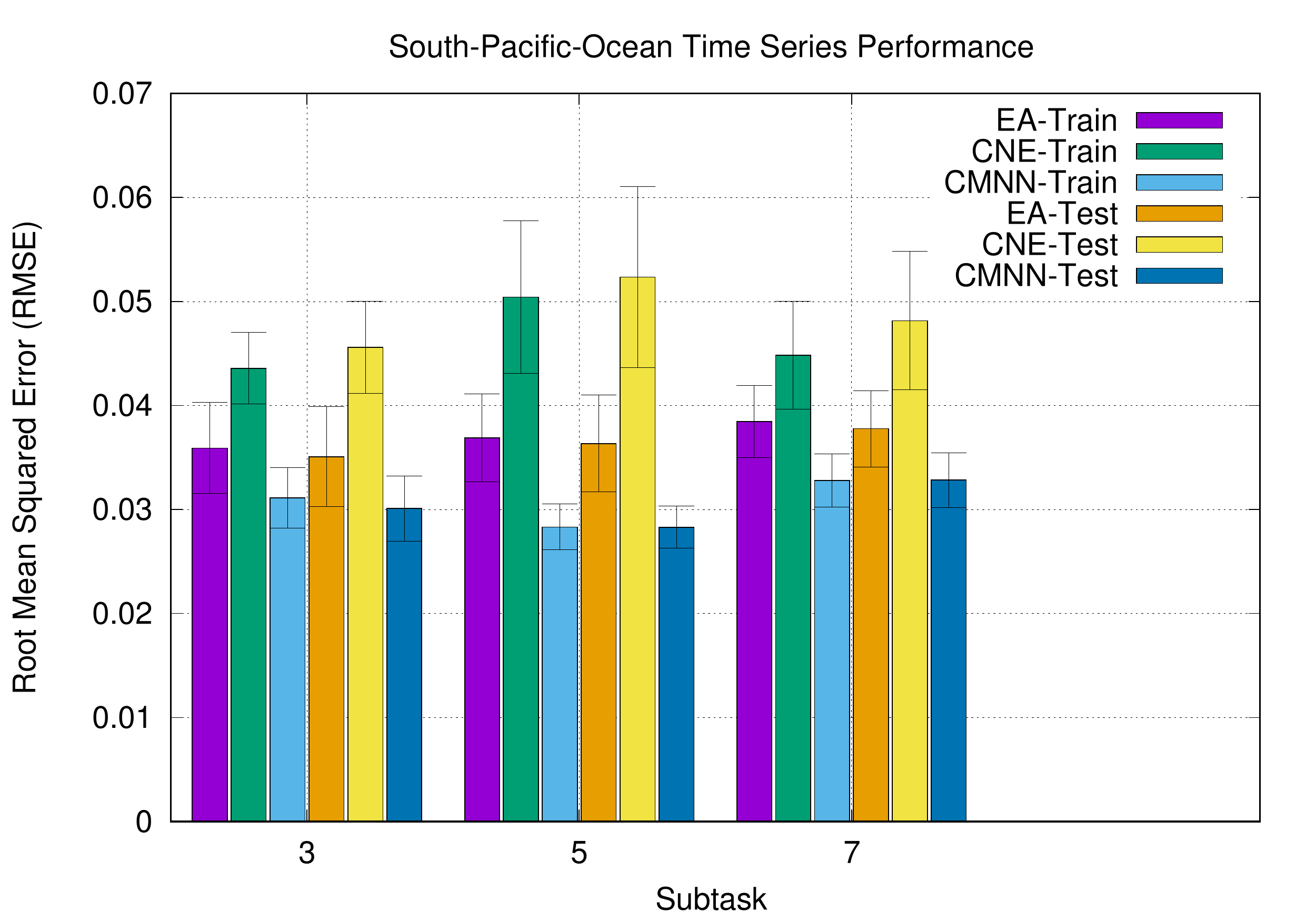}
\caption{Performance given by  EA, CNE, CMTL for South Pacific ocean}
\label{fig:sp}
\end{figure}

\begin{figure*}[ht!]
\centering
\includegraphics[width=100mm, angle= 
270]{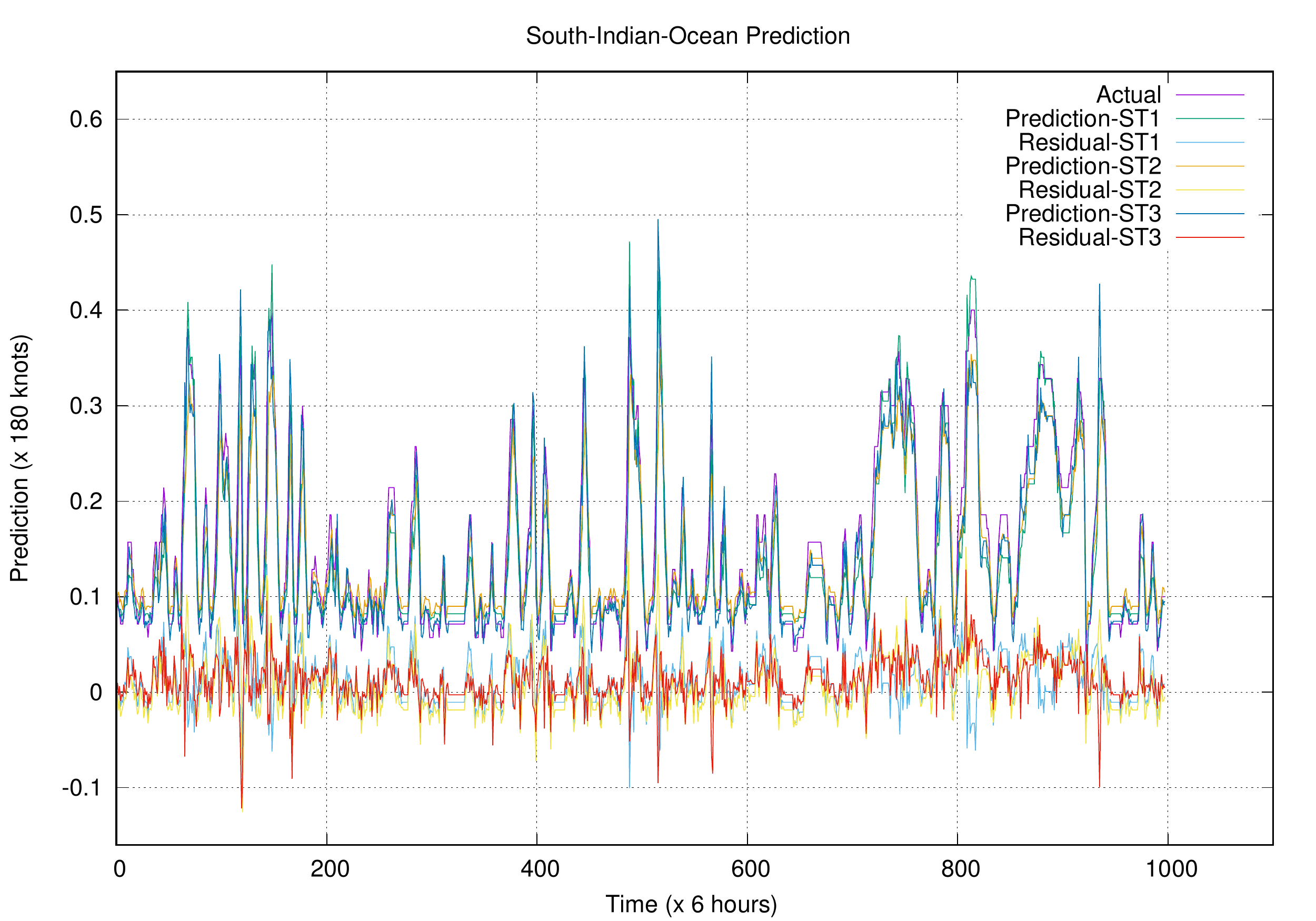}
\caption{ Typical prediction performance given by different CMTL subtasks (ST1, 
ST2, and ST3)   for the South Indian ocean}
\label{fig:iograph}
\end{figure*}

\begin{figure*}[ht!]
\centering
\includegraphics[width=100mm, angle= 
270]{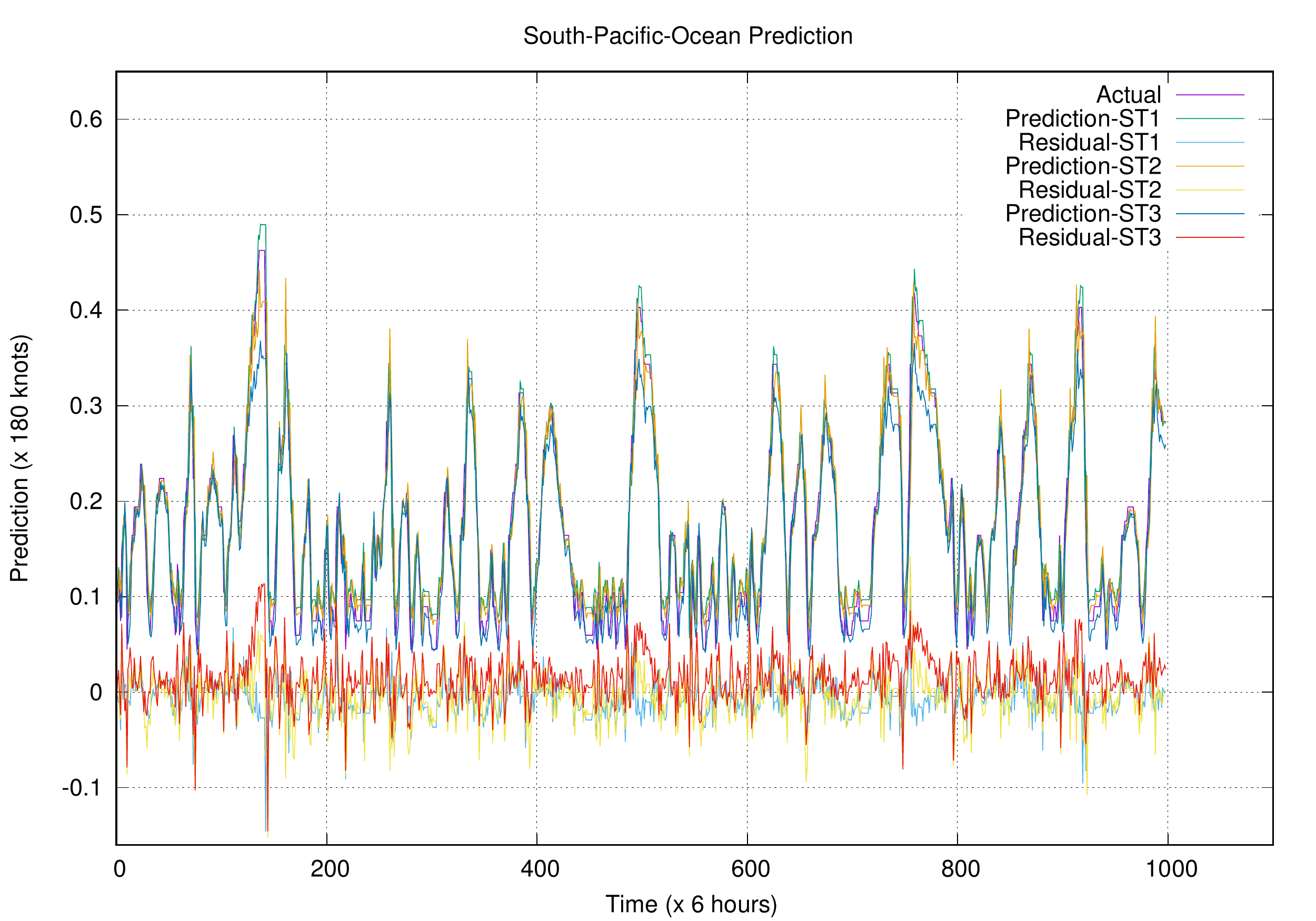}
\caption{Typical prediction performance given by different CMTL subtasks (ST1, 
ST2, and ST3)   for the  South Pacific 
ocean}
\label{fig:spgraph}
\end{figure*}

 We present the results for the performance of the given methods 
 on the 
two selected cyclone problems, which features  South Pacific and South Indian 
ocean as shown in Figure  \ref{fig:io} and Figure \ref{fig:sp}, 
respectively. Figures \ref{fig:spgraph} and \ref{fig:iograph} show the 
prediction performance of a typical experimental run.
 In the case of the South Pacific ocean, the results show that 
CMTL provides the 
best generalisation  performance when compared to CNE and EA. This is 
also observed for the South Indian ocean.

 \subsection{Discussion}
 
   The goal of the experiments was to evaluate if  CMTL can maintain quality in 
prediction performance when compared to conventional methods, while at the same 
time address dynamic time series problems. The results have shown that CMTL  not 
only addresses dynamic time series but is a  way to improve the performance if 
each of the subtasks in multi-task learning was decomposed and approached as 
single-task learning. The incremental learning in CMTL not only improves the 
prediction performance but also ensures modularity in the organisation of 
knowledge. Modularity is an important attribute for addressing dynamic 
prediction problems since groups of knowledge can be combined to make a decision 
when  the nature or complexity of the problem increases. Modularity is important 
for design of neural network in hardware  \cite{SurveyHardwareNN2010} as 
disruptions in certain synapse(s) can result in problems with the whole network 
which can be eliminated by persevering knowledge as modules \cite{Clune2012}.  

It is noteworthy that CMTL gives consistent performance even for the cases when 
the problem is harder as in the case of smaller or foundational subtasks that 
have minimal timespan.  Learning smaller networks could be harder since they 
have limited information about the past behaviour of the time series. The way 
the algorithm handles this issue  is with the refinement of the solutions (in a 
round-robin manner through coevolution) after it has been transferred to a 
larger network. When more information is presented as the subtask increases, 
CMTL tends to refine the knowledge in the smaller subtasks. In this way, the 
results show that the performance is consistent given the small subtasks and 
when they increase depending on the timespan.

 CMTL  can be seen 
as a flexible method for datasets that have different features of which some 
have properties so that they can be grouped together as subtasks.  Through 
multi-task learning, the overlapping features can be used as building blocks to 
learn 
the nature of the problem through the model at hand. Although feedforward 
neural networks have been used in CMTL, other   other neural network 
architectures and learning models can be used depending on the nature of 
the subtasks. In case of 
computer vision applications such as face recognition, the different subtasks 
can 
be different number of features; i.e.  the algorithm  can execute  face 
recognition based on minimal features from a set of features.

 CMTL could  also be viewed as   training cascaded 
networks  using a dynamic programming approach where each cascade defines a 
subtask. Although the 
depth 
of the cascading does not have a limit,  adding cascades 
could result in 
an exponential increase of training time.  The depth of the cascaded 
architecture would 
be dependent on the application problem. It depends on the time series under 
consideration and the level of inter-dependencies of the current and  the 
previous time steps. In principle, one should stop adding cascades when the 
prediction performance begins to deteriorate to a given threshold.  Therefore, 
for a given problem, there needs to be systematic approach that selects the 
number 
of subtasks for the cascaded architecture.

 We have experimentally tested robustness and scalability 
using the synthetic and real-world datasets that includes benchmark problems 
and an
application  that considers the prediction of wind-intensity in tropical 
cyclones. We provided comprehensive experimentation  for the algorithm 
convergence given a range of initial  conditions. These include different set 
of   
initialisation of the subpopulations in CMTL with multiple experimental runs  
along 
with further reporting of the mean and confidence interval. We evaluated the 
prediction capability given different instances of the timespan 
defined as subtasks in CMTL and compared the performance with standalone 
methods. The experimental design 
considered multiples experiment runs, different and distinct datasets, 
difference in
size of the datasets, and different sets of initialisation in the 
subpopulations. 
In this way, we have addressed robustness and convergence of the proposed 
method, experimentally.

 In the comparison of CMTL  with CNE, we observed that CMTL creates a 
higher time complexity since it has an additional step of transfer of solutions 
from the different subtasks encoded in the subpopulations. The time taken would 
increase exponentially as the number of subtasks increases. This would add to 
cost of utilising solutions from other subtasks given a fixed convergence 
criteria defined by the number of function evaluations. In case if the 
convergence 
criteria is defined by a minimum error or loss, it is likely that the solutions 
from the previous subtasks will help in faster convergence. In terms of 
scalability, we note that neuroevolution methods have limitations due to slow 
convergence of evolutionary algorithms. With help of gradient-based local 
search 
methods, convergence of  CMTL can be improved  via  memetic  algorithms where 
local refinement occurs during the evolution \cite{Chen2011}. There is scope in 
future work convergence proof used in standard evolutionary algorithms   
\cite{Eiben1990}    that could further be extended for multiple sub-populations 
in CMTL.

\section{Conclusions and Future Work}

We presented a novel algorithm that provides a synergy between coevolutionary 
algorithms  and 
multi-tasking for dynamic time series problems. CMTL can be used to train a 
model 
that can handle multiple 
timespan values that defines dynamic input features which provides dynamic  
prediction.   CMTL has been 
very beneficial  for  tropical 
cyclones where timely prediction needs to be made as soon as the event takes 
place.   The results 
show that CMTL  addresses  the 
problem of   dynamic time series and provides  a  robust way to improve the 
performance  when compared to related  methods.

It is important to understand how CMTL achieved better results for when 
compared 
to related  methods  given   the same neural network topology and data for 
respective subtasks.   CMTL can be seen as an incremental evolutionary learning 
method that features 
 subtasks as building blocks of knowledge. The larger  subtasks take 
advantage of 
knowledge 
gained  from learning the smaller subtasks.  Hence, there is diversity in 
incremental knowledge  
development  from the base subtask which seems to be beneficial for 
future subtasks. However, the reason why the base subtask produces better 
results 
when 
compared to  conventional learning   can be explored with further analysis 
during   learning. The   larger subtasks  with 
overlapping features covers the 
base subtask in a cascaded manner. Therefore,  larger subtasks can be seen as 
those that have 
additional 
features that guide  larger  network  with more hidden neurons  during 
training. 
 
 Finally, CMTL is a novel approach that provides synergy of  a 
wide range of fundamental methods that include, dynamic programming, 
reinforcement learning, multi-task learning,  co-evolutionary algorithms and  
neuro-evolution. This makes the CMTL useful for some of the   applications where 
the mentioned fundamental methods have been successful. Since reinforcement 
learning has been utilised in deep learning, the  notion of 
reuse of knowledge as building bocks by CMTL could be applicable in areas of 
deep learning. In future work, apart from feed-forward networks, the idea of 
dynamic time series prediction that employs transfer and multi-task learning 
could be extended to other other areas that have simpler model representation, 
such 
as autoregressive models. Furthermore, CMTL can be used for other  
problems that can be broken into multiple subtasks,  such as  multiple step 
ahead and multivariate time series  prediction.   Although the paper explored 
timespan for univariate time series, the approach could be extended to  
pattern classification problems that involves large scale features. It can be 
extended for heterogeneous pattern classification problems where the dataset 
contains 
samples that have missing values or features.  CMTL can also be 
extended for transfer 
learning problems that can include both heterogeneous and homogeneous domain 
adaptation cases. In case of tropical cyclones, a multivariate approach can be 
taken where  the different subtasks can be seen as   features that include 
cyclone tracks, seas surface temperature, and humidity. 
 \section*{References}
 \bibliographystyle{IEEEtran}
  
\bibliography{aicrg,rr,cyclone,mergedApril2015}

\begin{thebibliography}{100}
\providecommand{\url}[1]{#1}
\csname url@samestyle\endcsname
\providecommand{\newblock}{\relax}
\providecommand{\bibinfo}[2]{#2}
\providecommand{\BIBentrySTDinterwordspacing}{\spaceskip=0pt\relax}
\providecommand{\BIBentryALTinterwordstretchfactor}{4}
\providecommand{\BIBentryALTinterwordspacing}{\spaceskip=\fontdimen2\font plus
\BIBentryALTinterwordstretchfactor\fontdimen3\font minus
  \fontdimen4\font\relax}
\providecommand{\BIBforeignlanguage}[2]{{%
\expandafter\ifx\csname l@#1\endcsname\relax
\typeout{** WARNING: IEEEtran.bst: No hyphenation pattern has been}%
\typeout{** loaded for the language `#1'. Using the pattern for}%
\typeout{** the default language instead.}%
\else
\language=\csname l@#1\endcsname
\fi
#2}}
\providecommand{\BIBdecl}{\relax}
\BIBdecl

\bibitem{Mirikitani2010}
D.~Mirikitani and N.~Nikolaev, ``Recursive bayesian recurrent neural networks
  for time-series modeling,'' \emph{Neural Networks, IEEE Transactions on},
  vol.~21, no.~2, pp. 262 --274, Feb. 2010.

\bibitem{ArdalaniFarsa2010}
M.~Ardalani-Farsa and S.~Zolfaghari, ``Chaotic time series prediction with
  residual analysis method using hybrid {Elman-NARX} neural networks,''
  \emph{Neurocomputing}, vol.~73, no. 13-15, pp. 2540 -- 2553, 2010.

\bibitem{Teo2001}
K.~K. Teo, L.~Wang, and Z.~Lin, ``Wavelet packet multi-layer perceptron for
  chaotic time series prediction: Effects of weight initialization,'' in
  \emph{Proceedings of the International Conference on Computational
  Science-Part II}, ser. ICCS '01, 2001, pp. 310--317.

\bibitem{Gholipour2006}
A.~Gholipour, B.~N. Araabi, and C.~Lucas, ``Predicting chaotic time series
  using neural and neurofuzzy models: A comparative study,'' \emph{Neural
  Process. Lett.}, vol.~24, pp. 217--239, 2006.

\bibitem{RutaEnsembles2007}
D.~Ruta and B.~Gabrys, ``Neural network ensembles for time series prediction,''
  in \emph{2007 International Joint Conference on Neural Networks}, 2007, pp.
  1204--1209.

\bibitem{Lin2009}
C.-J. Lin, C.-H. Chen, and C.-T. Lin, ``A hybrid of cooperative particle swarm
  optimization and cultural algorithm for neural fuzzy networks and its
  prediction applications,'' \emph{Systems, Man, and Cybernetics, Part C:
  Applications and Reviews, IEEE Transactions on}, vol.~39, no.~1, pp. 55--68,
  Jan. 2009.

\bibitem{Chandra2012times}
R.~Chandra and M.~Zhang, ``Cooperative coevolution of {Elman} recurrent neural
  networks for chaotic time series prediction,'' \emph{Neurocomputing}, vol.
  186, pp. 116 -- 123, 2012.

\bibitem{ChandraTNNLS2015}
R.~Chandra, ``Competition and collaboration in cooperative coevolution of
  {Elman} recurrent neural networks for time-series prediction,'' \emph{Neural
  Networks and Learning Systems, IEEE Transactions on}, vol.~26, pp.
  3123--3136, 2015.

\bibitem{Takens1981}
F.~Takens, ``Detecting strange attractors in turbulence,'' in \emph{Dynamical
  Systems and Turbulence, Warwick 1980}, ser. Lecture Notes in Mathematics,
  1981, pp. 366--381.

\bibitem{NandC16FeatureS}
R.~Nand and R.~Chandra, ``Coevolutionary feature selection and reconstruction
  in neuro-evolution for time series prediction,'' in \emph{Artificial Life and
  Computational Intelligence - Second Australasian Conference, {ACALCI} 2016,
  Canberra, ACT, Australia, February 2-5, 2016, Proceedings}, 2016, pp.
  285--297.

\bibitem{ChandMO2014}
S.~Chand and R.~Chandra, ``Multi-objective cooperative coevolution of neural
  networks for time series prediction,'' in \emph{International Joint
  Conference on Neural Networks {(IJCNN)}}, Beijing, China, July 2014, pp.
  190--197.

\bibitem{Maus2011Embed}
A.~Maus and J.~Sprott, ``Neural network method for determining embedding
  dimension of a time series,'' \emph{Communications in Nonlinear Science and
  Numerical Simulation}, vol.~16, no.~8, pp. 3294 -- 3302, 2011.

\bibitem{HeatPotential2012}
M.~M. Ali, P.~S.~V. Jagadeesh, I.~I. Lin, and J.~Y. Hsu, ``A neural network
  approach to estimate tropical cyclone heat potential in the indian ocean,''
  \emph{IEEE Geoscience and Remote Sensing Letters}, vol.~9, no.~6, pp.
  1114--1117, Nov 2012.

\bibitem{Zjavka2016}
L.~Zjavka, ``Numerical weather prediction revisions using the locally trained
  differential polynomial network,'' \emph{Expert Systems with Applications},
  vol.~44, pp. 265 -- 274, 2016.

\bibitem{Winds2014NeuralNet}
B.~W. Stiles, R.~E. Danielson, W.~L. Poulsen, M.~J. Brennan,
  S.~Hristova-Veleva, T.~P. Shen, and A.~G. Fore, ``Optimized tropical cyclone
  winds from quikscat: A neural network approach,'' \emph{IEEE Transactions on
  Geoscience and Remote Sensing}, vol.~52, no.~11, pp. 7418--7434, Nov 2014.

\bibitem{RatneelIJCNN2016}
R.~Deo and R.~Chandra, ``Identification of minimal timespan problem for
  recurrent neural networks with application to cyclone wind-intensity
  prediction,'' in \emph{International Joint Conference on Neural Networks
  {(IJCNN)}}, Vancouver, Canada, July 2016, p. In Press.

\bibitem{taieb2015bias}
S.~B. Taieb and A.~F. Atiya, ``A bias and variance analysis for multistep-ahead
  time series forecasting,'' 2015.

\bibitem{chang2012reinforced}
L.-C. Chang, P.-A. Chen, and F.-J. Chang, ``Reinforced two-step-ahead weight
  adjustment technique for online training of recurrent neural networks,''
  \emph{Neural Networks and Learning Systems, IEEE Transactions on}, vol.~23,
  no.~8, pp. 1269--1278, 2012.

\bibitem{bone2002multi}
R.~Bon{\'e} and M.~Crucianu, ``Multi-step-ahead prediction with neural
  networks: a review,'' \emph{9emes rencontres internationales: Approches
  Connexionnistes en Sciences}, vol.~2, pp. 97--106, 2002.

\bibitem{Chakraborty1992}
K.~Chakraborty, K.~Mehrotra, C.~K. Mohan, and S.~Ranka, ``Forecasting the
  behavior of multivariate time series using neural networks,'' \emph{Neural
  Networks}, vol.~5, no.~6, pp. 961 -- 970, 1992.

\bibitem{Wang2016MVTS}
L.~Wang, Z.~Wang, and S.~Liu, ``An effective multivariate time series
  classification approach using echo state network and adaptive differential
  evolution algorithm,'' \emph{Expert Systems with Applications}, vol.~43, pp.
  237 -- 249, 2016.

\bibitem{Zhang2016}
S.~Zhang, ``Adaptive spectral estimation for nonstationary multivariate time
  series,'' \emph{Computational Statistics and Data Analysis}, vol. 103, pp.
  330 -- 349, 2016.

\bibitem{Caruana1997}
R.~Caruana, ``Multitask learning,'' \emph{Machine Learning}, vol.~28, no.~1,
  pp. 41--75, Jul. 1997.

\bibitem{evgeniou2005MT}
T.~Evgeniou, C.~A. Micchelli, and M.~Pontil, ``Learning multiple tasks with
  kernel methods,'' \emph{Journal of Machine Learning Research}, vol.~6, no.
  Apr, pp. 615--637, 2005.

\bibitem{MT-face-expZheng2016}
H.~Zheng, X.~Geng, D.~Tao, and Z.~Jin, ``A multi-task model for simultaneous
  face identification and facial expression recognition,''
  \emph{Neurocomputing}, vol. 171, pp. 515 -- 523, 2016.

\bibitem{MT-DeepNN2015}
T.~Zeng and S.~Ji, ``Deep convolutional neural networks for multi-instance
  multi-task learning,'' in \emph{Data Mining (ICDM), 2015 IEEE International
  Conference on}, Nov 2015, pp. 579--588.

\bibitem{Happel1994}
\BIBentryALTinterwordspacing
B.~L. Happel and J.~M. Murre, ``Design and evolution of modular neural network
  architectures,'' \emph{Neural Networks}, vol.~7, no. 6–7, pp. 985 -- 1004,
  1994, models of Neurodynamics and Behavior. [Online]. Available:
  \url{http://www.sciencedirect.com/science/article/pii/S0893608005801558}
\BIBentrySTDinterwordspacing

\bibitem{Clune2012}
J.~Clune, J.-B. Mouret, and H.~Lipson, ``The evolutionary origins of
  modularity,'' \emph{Proceedings of the Royal Society of London B: Biological
  Sciences}, vol. 280, no. 1755, 2013.

\bibitem{Ellefsen2015}
K.~O. Ellefsen, J.-B. Mouret, and J.~Clune, ``Neural modularity helps organisms
  evolve to learn new skills without forgetting old skills,'' \emph{PLoS Comput
  Biol}, vol.~11, no.~4, pp. 1--24, 04 2015.

\bibitem{neuro-DP1997}
J.~N. Tsitsiklis and B.~V. Roy, ``Neuro-dynamic programming overview and a case
  study in optimal stopping,'' in \emph{Decision and Control, 1997.,
  Proceedings of the 36th IEEE Conference on}, vol.~2, Dec 1997, pp. 1181--1186
  vol.2.

\bibitem{datadrivenDP2015}
X.~Fang, D.~Zheng, H.~He, and Z.~Ni, ``Data-driven heuristic dynamic
  programming with virtual reality,'' \emph{Neurocomputing}, vol. 166, pp. 244
  -- 255, 2015.

\bibitem{Potter_Jong1994}
M.~Potter and K.~De~Jong, ``A cooperative coevolutionary approach to function
  optimization,'' in \emph{Parallel Problem Solving from Nature — PPSN III},
  ser. Lecture Notes in Computer Science, Y.~Davidor, H.-P. Schwefel, and
  R.~Männer, Eds.\hskip 1em plus 0.5em minus 0.4em\relax Springer Berlin
  Heidelberg, 1994, vol. 866, pp. 249--257.

\bibitem{Potter2000}
M.~A. Potter and K.~A. De~Jong, ``Cooperative coevolution: An architecture for
  evolving coadapted subcomponents,'' \emph{Evol. Comput.}, vol.~8, pp. 1--29,
  2000.

\bibitem{Gupta2016TEC}
A.~Gupta, Y.~S. Ong, and L.~Feng, ``Multifactorial evolution: Toward
  evolutionary multitasking,'' \emph{{IEEE} Trans. Evolutionary Computation},
  vol.~20, no.~3, pp. 343--357, 2016.

\bibitem{YewSoon2016}
Y.~S. Ong and A.~Gupta, ``Evolutionary multitasking: {A} computer science view
  of cognitive multitasking,'' \emph{Cognitive Computation}, vol.~8, no.~2, pp.
  125--142, 2016.

\bibitem{Chandra2017NPL}
\BIBentryALTinterwordspacing
R.~Chandra, A.~Gupta, Y.~S. Ong, and C.~K. Goh, ``Evolutionary multi-task
  learning for modular knowledge representation in neural networks,''
  \emph{Neural Processing Letters}, 2018. [Online]. Available:
  \url{https://doi.org/10.1007/s11063-017-9718-z}
\BIBentrySTDinterwordspacing

\bibitem{ChandraICONIP2017}
R.~Chandra, ``Dynamic cyclone wind-intensity prediction using co-evolutionary
  multi-task learning,'' in \emph{Neural Information Processing}, D.~Liu,
  S.~Xie, Y.~Li, D.~Zhao, and E.-S.~M. El-Alfy, Eds., 2017, pp. 618--627.

\bibitem{Ando2005}
\BIBentryALTinterwordspacing
R.~K. Ando and T.~Zhang, ``A framework for learning predictive structures from
  multiple tasks and unlabeled data,'' \emph{J. Mach. Learn. Res.}, vol.~6, pp.
  1817--1853, Dec. 2005. [Online]. Available:
  \url{http://dl.acm.org/citation.cfm?id=1046920.1194905}
\BIBentrySTDinterwordspacing

\bibitem{NIPS2008_3499}
\BIBentryALTinterwordspacing
L.~Jacob, J.~philippe Vert, and F.~R. Bach, ``Clustered multi-task learning: A
  convex formulation,'' in \emph{Advances in Neural Information Processing
  Systems 21}, D.~Koller, D.~Schuurmans, Y.~Bengio, and L.~Bottou, Eds.\hskip
  1em plus 0.5em minus 0.4em\relax Curran Associates, Inc., 2009, pp. 745--752.
  [Online]. Available:
  \url{http://papers.nips.cc/paper/3499-clustered-multi-task-learning-a-convex-formulation.pdf}
\BIBentrySTDinterwordspacing

\bibitem{Chen2009ICML}
\BIBentryALTinterwordspacing
J.~Chen, L.~Tang, J.~Liu, and J.~Ye, ``A convex formulation for learning shared
  structures from multiple tasks,'' in \emph{Proceedings of the 26th Annual
  International Conference on Machine Learning}, ser. ICML '09.\hskip 1em plus
  0.5em minus 0.4em\relax New York, NY, USA: ACM, 2009, pp. 137--144. [Online].
  Available: \url{http://doi.acm.org/10.1145/1553374.1553392}
\BIBentrySTDinterwordspacing

\bibitem{ZhouNIPS2011}
\BIBentryALTinterwordspacing
J.~Zhou, J.~Chen, and J.~Ye, ``Clustered multi-task learning via alternating
  structure optimization,'' in \emph{Advances in Neural Information Processing
  Systems 24}, J.~Shawe-Taylor, R.~S. Zemel, P.~L. Bartlett, F.~Pereira, and
  K.~Q. Weinberger, Eds.\hskip 1em plus 0.5em minus 0.4em\relax Curran
  Associates, Inc., 2011, pp. 702--710. [Online]. Available:
  \url{http://papers.nips.cc/paper/4292-clustered-multi-task-learning-via-alternating-structure-optimization.pdf}
\BIBentrySTDinterwordspacing

\bibitem{Zhang2010TML}
\BIBentryALTinterwordspacing
Y.~Zhang and D.-Y. Yeung, ``Transfer metric learning by learning task
  relationships,'' in \emph{Proceedings of the 16th ACM SIGKDD International
  Conference on Knowledge Discovery and Data Mining}, ser. KDD '10.\hskip 1em
  plus 0.5em minus 0.4em\relax New York, NY, USA: ACM, 2010, pp. 1199--1208.
  [Online]. Available: \url{http://doi.acm.org/10.1145/1835804.1835954}
\BIBentrySTDinterwordspacing

\bibitem{Bakker2003}
\BIBentryALTinterwordspacing
B.~Bakker and T.~Heskes, ``Task clustering and gating for bayesian multitask
  learning,'' \emph{J. Mach. Learn. Res.}, vol.~4, pp. 83--99, Dec. 2003.
  [Online]. Available: \url{http://dx.doi.org/10.1162/153244304322765658}
\BIBentrySTDinterwordspacing

\bibitem{Zhong2016}
\BIBentryALTinterwordspacing
S.~Zhong, J.~Pu, Y.-G. Jiang, R.~Feng, and X.~Xue, ``Flexible multi-task
  learning with latent task grouping,'' \emph{Neurocomputing}, vol. 189, pp.
  179 -- 188, 2016. [Online]. Available:
  \url{http://www.sciencedirect.com/science/article/pii/S0925231216000035}
\BIBentrySTDinterwordspacing

\bibitem{Tang2015}
\BIBentryALTinterwordspacing
X.~Tang, Q.~Miao, Y.~Quan, J.~Tang, and K.~Deng, ``Predicting individual
  retweet behavior by user similarity: A multi-task learning approach,''
  \emph{Knowledge-Based Systems}, vol.~89, pp. 681 -- 688, 2015. [Online].
  Available:
  \url{http://www.sciencedirect.com/science/article/pii/S0950705115003470}
\BIBentrySTDinterwordspacing

\bibitem{Liu2016NC}
\BIBentryALTinterwordspacing
A.~Liu, Y.~Lu, W.~Nie, Y.~Su, and Z.~Yang, ``Hep-2 cells classification via
  clustered multi-task learning,'' \emph{Neurocomputing}, vol. 195, pp. 195 --
  201, 2016, learning for Medical Imaging. [Online]. Available:
  \url{http://www.sciencedirect.com/science/article/pii/S0925231216001235}
\BIBentrySTDinterwordspacing

\bibitem{Qin2016Kinship}
\BIBentryALTinterwordspacing
X.~Qin, X.~Tan, and S.~Chen, ``Mixed bi-subject kinship verification via
  multi-view multi-task learning,'' \emph{Neurocomputing}, pp.~--, 2016.
  [Online]. Available:
  \url{http://www.sciencedirect.com/science/article/pii/S0925231216306658}
\BIBentrySTDinterwordspacing

\bibitem{Zhang2015OTrack}
\BIBentryALTinterwordspacing
S.~Zhang, Y.~Sui, S.~Zhao, X.~Yu, and L.~Zhang, ``Multi-local-task learning
  with global regularization for object tracking,'' \emph{Pattern Recognition},
  vol.~48, no.~12, pp. 3881 -- 3894, 2015. [Online]. Available:
  \url{http://www.sciencedirect.com/science/article/pii/S0031320315002265}
\BIBentrySTDinterwordspacing

\bibitem{AngelineGNARL1994}
P.~Angeline, G.~Saunders, and J.~Pollack, ``An evolutionary algorithm that
  constructs recurrent neural networks,'' \emph{Neural Networks, IEEE
  Transactions on}, vol.~5, no.~1, pp. 54 --65, jan 1994.

\bibitem{SANE1997}
\BIBentryALTinterwordspacing
D.~E. Moriarty and R.~Miikkulainen, ``Forming neural networks through efficient
  and adaptive coevolution,'' \emph{Evolutionary Computation}, vol.~5, no.~4,
  pp. 373--399, 1997. [Online]. Available:
  \url{http://www.mitpressjournals.org/doi/abs/10.1162/evco.1997.5.4.373}
\BIBentrySTDinterwordspacing

\bibitem{StanleyNEAT2002}
K.~O. Stanley and R.~Miikkulainen, ``Evolving neural networks through
  augmenting topologies,'' \emph{Evolutionary Computation}, vol.~10, no.~2, pp.
  99--127, 2002.

\bibitem{Gomez_Schmidhuber2008}
F.~Gomez, J.~Schmidhuber, and R.~Miikkulainen, ``Accelerated neural evolution
  through cooperatively coevolved synapses,'' \emph{J. Mach. Learn. Res.},
  vol.~9, pp. 937--965, 2008.

\bibitem{HeidrichMeisner2009}
\BIBentryALTinterwordspacing
V.~Heidrich-Meisner and C.~Igel, ``Neuroevolution strategies for episodic
  reinforcement learning,'' \emph{Journal of Algorithms}, vol.~64, no.~4, pp.
  152 -- 168, 2009, special Issue: Reinforcement Learning. [Online]. Available:
  \url{http://www.sciencedirect.com/science/article/B6WH3-4W7RY8J-3/2/22f7075bc25dab10a8ff3714e2fee303}
\BIBentrySTDinterwordspacing

\bibitem{MobNet2002}
N.~García-Pedrajas, C.~Hervas-Martinez, and J.~Munoz-Perez, ``Multi-objective
  cooperative coevolution of artificial neural networks (multi-objective
  cooperative networks),'' \emph{Neural Networks}, vol.~15, pp. 1259--1278,
  2002.

\bibitem{Chandra2012sep}
R.~Chandra, M.~Frean, and M.~Zhang, ``On the issue of separability for problem
  decomposition in cooperative neuro-evolution,'' \emph{Neurocomputing},
  vol.~87, pp. 33--40, 2012.

\bibitem{chandra2010}
------, ``An encoding scheme for cooperative coevolutionary neural networks,''
  in \emph{23rd Australian Joint Conference on Artificial Intelligence}, ser.
  Lecture Notes in Artificial Intelligence.\hskip 1em plus 0.5em minus
  0.4em\relax Adelaide, Australia: Springer-Verlag, 2010, pp. 253--262.

\bibitem{ChandraNSPRNN2011}
R.~Chandra, M.~Frean, M.~Zhang, and C.~W. Omlin, ``Encoding subcomponents in
  cooperative co-evolutionary recurrent neural networks,''
  \emph{Neurocomputing}, vol.~74, no.~17, pp. 3223 -- 3234, 2011.

\bibitem{ESP_Gomez1997}
F.~Gomez and R.~Mikkulainen, ``Incremental evolution of complex general
  behavior,'' \emph{Adapt. Behav.}, vol.~5, no. 3-4, pp. 317--342, 1997.

\bibitem{GomezPhD2003}
F.~J. Gomez, ``Robust non-linear control through neuroevolution,'' PhD Thesis,
  Department of Computer Science, The University of Texas at Austin, Technical
  Report AI-TR-03-303, 2003.

\bibitem{howard1966dynamic}
R.~A. Howard, ``Dynamic programming,'' \emph{Management Science}, vol.~12,
  no.~5, pp. 317--348, 1966.

\bibitem{held1962dynamic}
M.~Held and R.~M. Karp, ``A dynamic programming approach to sequencing
  problems,'' \emph{Journal of the Society for Industrial and Applied
  Mathematics}, vol.~10, no.~1, pp. 196--210, 1962.

\bibitem{sakoe1978dynamic}
H.~Sakoe and S.~Chiba, ``Dynamic programming algorithm optimization for spoken
  word recognition,'' \emph{IEEE transactions on acoustics, speech, and signal
  processing}, vol.~26, no.~1, pp. 43--49, 1978.

\bibitem{amini1990using}
A.~A. Amini, T.~E. Weymouth, and R.~C. Jain, ``Using dynamic programming for
  solving variational problems in vision,'' \emph{IEEE Transactions on pattern
  analysis and machine intelligence}, vol.~12, no.~9, pp. 855--867, 1990.

\bibitem{sutton1992introduction}
R.~S. Sutton, ``Introduction: The challenge of reinforcement learning,'' in
  \emph{Reinforcement Learning}.\hskip 1em plus 0.5em minus 0.4em\relax
  Springer, 1992, pp. 1--3.

\bibitem{kaelbling1996reinforcement}
L.~P. Kaelbling, M.~L. Littman, and A.~W. Moore, ``Reinforcement learning: A
  survey,'' \emph{Journal of artificial intelligence research}, vol.~4, pp.
  237--285, 1996.

\bibitem{bertsekas1995dynamic}
D.~P. Bertsekas, D.~P. Bertsekas, D.~P. Bertsekas, and D.~P. Bertsekas,
  \emph{Dynamic programming and optimal control}.\hskip 1em plus 0.5em minus
  0.4em\relax Athena scientific Belmont, MA, 1995, vol.~1, no.~2.

\bibitem{bertsekas1995neuro}
D.~P. Bertsekas and J.~N. Tsitsiklis, ``Neuro-dynamic programming: an
  overview,'' in \emph{Decision and Control, 1995., Proceedings of the 34th
  IEEE Conference on}, vol.~1.\hskip 1em plus 0.5em minus 0.4em\relax IEEE,
  1995, pp. 560--564.

\bibitem{mnih2015human}
V.~Mnih, K.~Kavukcuoglu, D.~Silver, A.~A. Rusu, J.~Veness, M.~G. Bellemare,
  A.~Graves, M.~Riedmiller, A.~K. Fidjeland, G.~Ostrovski \emph{et~al.},
  ``Human-level control through deep reinforcement learning,'' \emph{Nature},
  vol. 518, no. 7540, p. 529, 2015.

\bibitem{moriarty1999evolutionary}
D.~E. Moriarty, A.~C. Schultz, and J.~J. Grefenstette, ``Evolutionary
  algorithms for reinforcement learning,'' \emph{Journal of Artifical
  Intelligence Research {JAIR}}, vol.~11, pp. 241 -- 276, 1999.

\bibitem{richards1998evolving}
N.~Richards, D.~E. Moriarty, and R.~Miikkulainen, ``Evolving neural networks to
  play go,'' \emph{Applied Intelligence}, vol.~8, no.~1, pp. 85--96, 1998.

\bibitem{bennett2006interplay}
K.~P. Bennett and E.~Parrado-Hern{\'a}ndez, ``The interplay of optimization and
  machine learning research,'' \emph{Journal of Machine Learning Research},
  vol.~7, no. Jul, pp. 1265--1281, 2006.

\bibitem{guillory2009active}
A.~Guillory, E.~Chastain, and J.~Bilmes, ``Active learning as non-convex
  optimization,'' in \emph{Artificial Intelligence and Statistics}, 2009, pp.
  201--208.

\bibitem{Koskela96}
T.~Koskela, M.~Lehtokangas, J.~Saarinen, and K.~Kaski, ``Time series prediction
  with multilayer perceptron, {FIR and Elman} neural networks,'' in \emph{In
  Proceedings of the World Congress on Neural Networks}, San Diego, CA, USA,
  1996, pp. 491--496.

\bibitem{sandya2013feature}
H.~Sandya, P.~Hemanth~Kumar, and S.~B. Patil, ``Feature extraction,
  classification and forecasting of time series signal using fuzzy and garch
  techniques,'' in \emph{Research \& Technology in the Coming Decades (CRT
  2013), National Conference on Challenges in}.\hskip 1em plus 0.5em minus
  0.4em\relax IET, 2013, pp. 1--7.

\bibitem{zhang2013iterated}
L.~Zhang, W.-D. Zhou, P.-C. Chang, J.-W. Yang, and F.-Z. Li, ``Iterated time
  series prediction with multiple support vector regression models,''
  \emph{Neurocomputing}, vol.~99, pp. 411--422, 2013.

\bibitem{ben2012recursive}
S.~Ben~Taieb and R.~Hyndman, ``Recursive and direct multi-step forecasting: the
  best of both worlds,'' Monash University, Department of Econometrics and
  Business Statistics, Tech. Rep., 2012.

\bibitem{Grigorievskiy2014}
A.~Grigorievskiy, Y.~Miche, A.-M. Ventelä, E.~Séverin, and A.~Lendasse,
  ``Long-term time series prediction using op-elm,'' \emph{Neural Networks},
  vol.~51, pp. 50 -- 56, 2014.

\bibitem{Yin2016}
\BIBentryALTinterwordspacing
Y.~Yin and P.~Shang, ``Forecasting traffic time series with multivariate
  predicting method,'' \emph{Applied Mathematics and Computation}, vol. 291,
  pp. 266 -- 278, 2016. [Online]. Available:
  \url{http://www.sciencedirect.com/science/article/pii/S0096300316304477}
\BIBentrySTDinterwordspacing

\bibitem{ChandraDayalIJCNN2015}
R.~Chandra, K.~Dayal, and N.~Rollings, ``Application of cooperative
  neuro-evolution of {Elman} recurrent networks for a two-dimensional cyclone
  track prediction for the {South Pacific} region,'' in \emph{International
  Joint Conference on Neural Networks {(IJCNN)}}, Killarney, Ireland, July
  2015, pp. 721--728.

\bibitem{Chayama2016}
\BIBentryALTinterwordspacing
M.~Chayama and Y.~Hirata, ``When univariate model-free time series prediction
  is better than multivariate,'' \emph{Physics Letters A}, vol. 380, no.
  31–32, pp. 2359 -- 2365, 2016. [Online]. Available:
  \url{http://www.sciencedirect.com/science/article/pii/S0375960116302195}
\BIBentrySTDinterwordspacing

\bibitem{Wu2014missingdate}
\BIBentryALTinterwordspacing
X.~Wu, Y.~Wang, J.~Mao, Z.~Du, and C.~Li, ``Multi-step prediction of time
  series with random missing data,'' \emph{Applied Mathematical Modelling},
  vol.~38, no.~14, pp. 3512 -- 3522, 2014. [Online]. Available:
  \url{http://www.sciencedirect.com/science/article/pii/S0307904X13007658}
\BIBentrySTDinterwordspacing

\bibitem{smith1989extreme}
R.~L. Smith, ``Extreme value analysis of environmental time series: an
  application to trend detection in ground-level ozone,'' \emph{Statistical
  Science}, pp. 367--377, 1989.

\bibitem{Durso2017}
P.~D'Urso, E.~A. Maharaj, and A.~M. Alonso, ``Fuzzy clustering of time series
  using extremes,'' \emph{Fuzzy Sets and Systems}, vol. 318, pp. 56 -- 79,
  2017, theme: Clustering and Image Processing.

\bibitem{Chouikhi2017}
\BIBentryALTinterwordspacing
N.~Chouikhi, B.~Ammar, N.~Rokbani, and A.~M. Alimi, ``Pso-based analysis of
  echo state network parameters for time series forecasting,'' \emph{Applied
  Soft Computing}, vol.~55, pp. 211 -- 225, 2017. [Online]. Available:
  \url{http://www.sciencedirect.com/science/article/pii/S1568494617300649}
\BIBentrySTDinterwordspacing

\bibitem{frazier2004}
C.~Frazier and K.~Kockelman, ``Chaos theory and transportation systems:
  Instructive example,'' \emph{Transportation Research Record: Journal of the
  Transportation Research Board}, vol.~20, pp. 9--17, 2004.

\bibitem{Calvo2000}
R.~A. Calvo, H.~D. Navone, and H.~A. Ceccatto, \emph{Neural Network Analysis of
  Time Series: Applications to Climatic Data}.\hskip 1em plus 0.5em minus
  0.4em\relax Berlin, Heidelberg: Springer Berlin Heidelberg, 2000, pp. 7--16.

\bibitem{YuanfeiBPNNCyclone2011}
Y.~Wang, W.~Zhang, and W.~Fu, ``Back propogation(bp)-neural network for
  tropical cyclone track forecast,'' in \emph{Geoinformatics, 2011 19th
  International Conference on}, June 2011, pp. 1--4.

\bibitem{JTWC}
(2015) {JTWC} tropical cyclone best track data site.

\bibitem{Mackey1977}
M.~Mackey and L.~Glass, ``Oscillation and chaos in physiological control
  systems,'' \emph{Science}, vol. 197, no. 4300, pp. 287--289, 1977.

\bibitem{lorenz1963}
E.~Lorenz, ``Deterministic non-periodic flows,'' \emph{Journal of Atmospheric
  Science}, vol.~20, pp. 267 -- 285, 1963.

\bibitem{Henon1976}
M.~H{\'e}non, ``A two-dimensional mapping with a strange attractor,''
  \emph{Communications in Mathematical Physics}, vol.~50, no.~1, pp. 69--77,
  1976.

\bibitem{rossler}
O.~Rössler, ``An equation for continuous chaos,'' \emph{Physics Letters A},
  vol.~57, no.~5, pp. 397 -- 398, 1976.

\bibitem{Sunspot2001}
S.~S., ``Solar cycle forecasting: A nonlinear dynamics approach,''
  \emph{Astronomy and Astrophysics}, vol. 377, pp. 312--320, 2001.

\bibitem{timeDataSet}
\BIBentryALTinterwordspacing
``{NASDAQ Exchange Daily: 1970-2010 Open, Close, High, Low and Volume},''
  accessed: 02-02-2015. [Online]. Available:
  \url{http://www.nasdaq.com/symbol/aciw/stock-chart}
\BIBentrySTDinterwordspacing

\bibitem{lazerdata}
\BIBentryALTinterwordspacing
``{Sante Fe Competition Data}.'' [Online]. Available:
  \url{http://www-psych.stanford.edu/~andreas/Time-Series/SantaFe.html, note =
  {Accessed: 30-06-2016}}
\BIBentrySTDinterwordspacing

\bibitem{HansenMK03}
\BIBentryALTinterwordspacing
N.~Hansen, S.~D. M{\"{u}}ller, and P.~Koumoutsakos, ``Reducing the time
  complexity of the derandomized evolution strategy with covariance matrix
  adaptation {(CMA-ES)},'' \emph{Evolutionary Computation}, vol.~11, no.~1, pp.
  1--18, 2003. [Online]. Available:
  \url{http://dx.doi.org/10.1162/106365603321828970}
\BIBentrySTDinterwordspacing

\bibitem{chandra2015competition}
\BIBentryALTinterwordspacing
R.~Chandra, ``Competition and collaboration in cooperative coevolution of elman
  recurrent neural networks for time-series prediction,'' \emph{{IEEE} Trans.
  Neural Netw. Learning Syst.}, vol.~26, no.~12, pp. 3123--3136, 2015.
  [Online]. Available: \url{http://dx.doi.org/10.1109/TNNLS.2015.2404823}
\BIBentrySTDinterwordspacing

\bibitem{SurveyHardwareNN2010}
J.~Misra and I.~Saha, ``Artificial neural networks in hardware: A survey of two
  decades of progress,'' \emph{Neurocomputing}, vol.~74, no. 1–3, pp. 239 --
  255, 2010, artificial Brains.

\bibitem{Chen2011}
X.~Chen, Y.~S. Ong, M.~H. Lim, and K.~C. Tan, ``A multi-facet survey on memetic
  computation,'' \emph{IEEE Transactions on Evolutionary Computation}, vol.~15,
  no.~5, pp. 591--607, 2011.

\bibitem{Eiben1990}
A.~E. Eiben, E.~H.~L. Aarts, and K.~M.~v. Hee, ``Global convergence of genetic
  algorithms: A markov chain analysis,'' in \emph{Proceedings of the 1st
  Workshop on Parallel Problem Solving from Nature}, ser. PPSN, 1991, pp.
  4--12.

\end{thebibliography}

\end{document}